\pdfoutput=1 
\documentclass[sigconf]{acmart}

\usepackage{amsthm}
\usepackage{enumitem}
\usepackage{caption}
\theoremstyle{definition}
\newtheorem{definition}{Definition}[section]
\AtBeginDocument{%
  \providecommand\BibTeX{{%
    \normalfont B\kern-0.5em{\scshape i\kern-0.25em b}\kern-0.8em\TeX}}}
\copyrightyear{2020}
\acmYear{2020}
\setcopyright{acmlicensed}
\acmConference[KDD '20]{Proceedings of the 26th ACM SIGKDD Conference on Knowledge Discovery and Data Mining USB
Stick}{August 23--27, 2020}{Virtual Event, USA}
\acmBooktitle{Proceedings of the 26th ACM SIGKDD Conference on Knowledge
Discovery and Data Mining USB Stick (KDD '20), August 23--27, 2020, Virtual
Event, USA}
\acmPrice{15.00}
\acmDOI{10.1145/3394486.3403275}
\acmISBN{978-1-4503-7998-4/20/08}

\begin{document}

\title{TIMME: Twitter Ideology-detection via Multi-task Multi-relational Embedding}


\author{Zhiping Xiao}
\affiliation{%
  \institution{CS Department, UCLA}
  \city{Los Angeles}
  \state{CA}
  \country{United States}
}
\email{patricia.xiao@cs.ucla.edu}

\author{Weiping Song}
\affiliation{%
  \institution{CS Department, School of EECS, Peking University}
  \city{Beijing}
  \country{China}
}
\email{weiping.song@pku.edu.cn}

\author{Haoyan Xu}
\affiliation{%
 \institution{College of Control Science and Engineering, Zhejiang University}
 \city{Hangzhou}
 \country{China}}
\email{3160104027@zju.edu.cn}

\author{Zhicheng Ren}
\affiliation{%
  \institution{CS Department, UCLA}
  \city{Los Angeles}
  \state{CA}
  \country{United States}
}
\email{franklinnwren@g.ucla.edu}

\author{Yizhou Sun}
\affiliation{%
  \institution{CS Department, UCLA}
  \city{Los Angeles}
  \state{CA}
  \country{United States}
}
\email{yzsun@cs.ucla.edu}


\begin{abstract}
    We aim at solving the problem of predicting people's ideology, or political tendency. We estimate it by using Twitter data, and formalize it as a classification problem.
    Ideology-detection has long been a challenging yet important problem. Certain groups, such as the policy makers, rely on it to make wise decisions. 
    Back in the old days when labor-intensive survey-studies were needed to collect public opinions, analyzing ordinary citizens' political tendencies was uneasy.
    The rise of social medias, such as Twitter, has enabled us to gather ordinary citizen's data easily.
    However, the incompleteness of the labels and the features in social network datasets is tricky, not to mention the enormous data size and the heterogeneousity. The data differ dramatically from many commonly-used datasets, thus brings unique challenges.
    In our work, first we built our own datasets from Twitter. Next, we proposed \textbf{TIMME}, a multi-task multi-relational embedding model, that works efficiently on sparsely-labeled heterogeneous real-world dataset. It could also handle the incompleteness of the input features.
    Experimental results showed that \textbf{TIMME} is overall better than the state-of-the-art models for ideology detection on Twitter.
    Our findings include: links can lead to good classification outcomes without text; conservative voice is under-represented on Twitter; \textit{follow} is the most important relation to predict ideology; \textit{retweet} and \textit{mention} enhance a higher chance of \textit{like}, etc. Last but not least, \textbf{TIMME} could be extended to other datasets and tasks in theory.
  
\end{abstract}

\ccsdesc[500]{Computing methodologies~Multi-task learning}
\ccsdesc[500]{Computing methodologies~Neural networks}

\keywords{multi-task learning, ideology detection, heterogeneous information network, social network analysis, graph convolutional networks}

\maketitle

\section{Introduction}\label{sec:intro}

\begin{figure}
\centering
\includegraphics[width=0.9\linewidth]{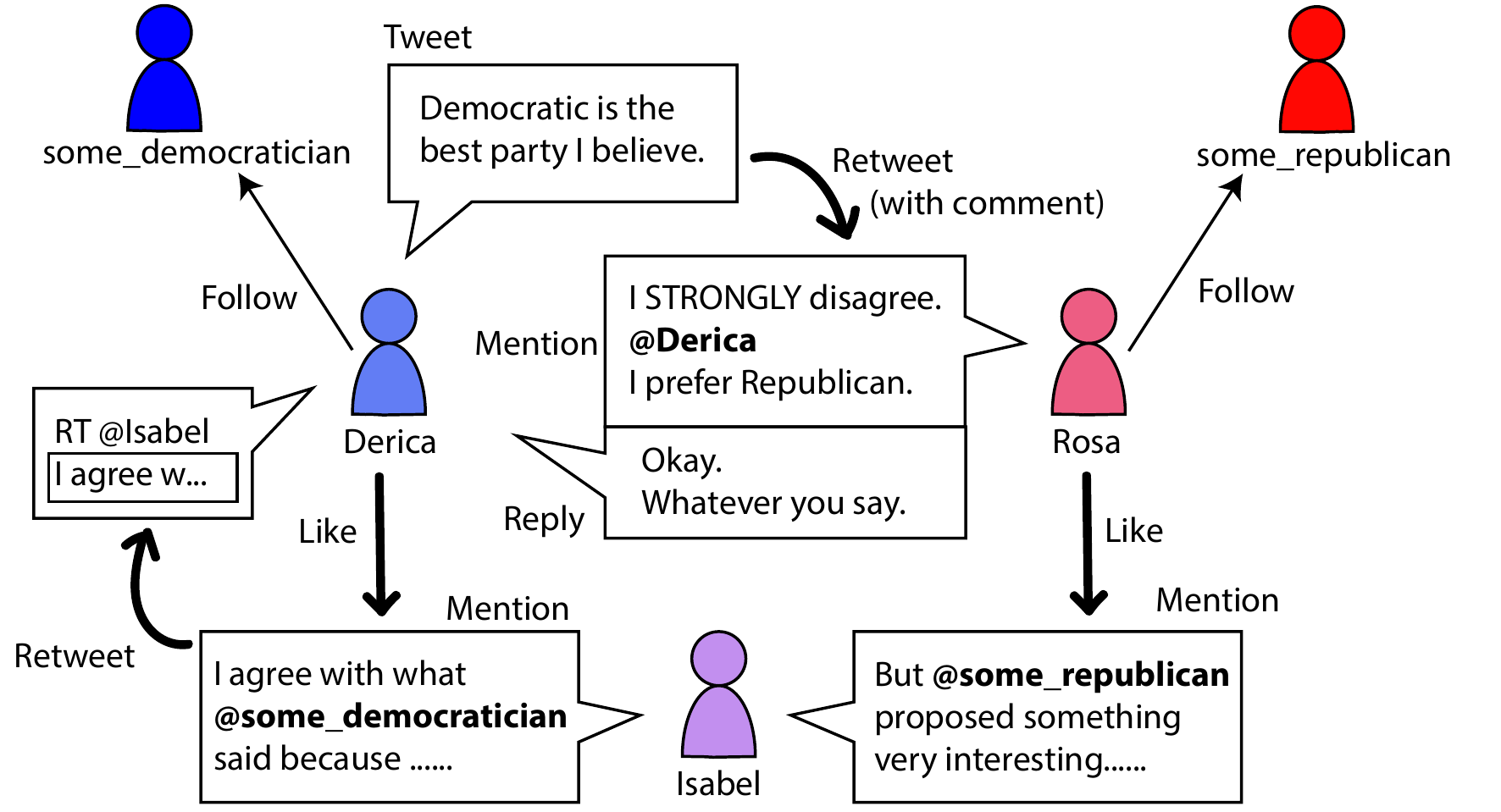}
\caption{An example of different relation types on Twitter. Derica is on liberal (left) side while Rosa is on the conservative (right) side. Isabel does not have significant tendency.}
\label{fig:illustrate}
\vspace{-3mm}
\end{figure}

Studies on ideology never fails to attract people's interests. Ideology here refers to the political stance or tendency of people, often reflected as left- or right-leaning. Measuring the politicians' ideology helps predict some important decisions' final outcomes, but it does not provide more insights into ordinary citizens' views, which are also of decisive significance.
Decades ago, social scientists have already started using probabilistic models to study the voting behaviors of the politicians. But seldom did they study the mass population's opinions, for the survey-based study is extremely labor-intensive and hard-to-scale \cite{achen1975mass,pollock2015populism}.
The booming development of social networks in the recent years shed light on detecting ordinary people's ideology. In social networks, people are more relaxed  than in an offline interview, and behave naturally. Social networks, in return, has shaped people's habits, giving rise to opinion leaders, encouraging youngsters' political involvement \cite{park2013does}.

Most existing approaches of ideology detection on social networks focus on text \cite{iyyer2014political,kannangara2018mining,chen2017opinion,johnson2016identifying,conover2011predicting}. Most of their methodologies based on probabilistic models, following the long-lasting tradition started by social scientists. Some others~\cite{baly2019multi,gu2016ideology,kannangara2018mining,preoctiuc2017beyond} noticed the advantages of neural networks, but seldom do they focus on links. We will show that the social-network links' contribution to ideology detection has been under-estimated.

An intuitive explanation of how links could be telling is illustrated in Figure \ref{fig:illustrate}. Different types of links come into being for different reasons. We have five relation types among users on Twitter today: \textit{follow, retweet, reply, mention, like}, and the relations affect each other. For instance, after Rosa \textit{retweet} from Derica and \textit{mention} her, Derica \textit{reply} to her; when Isabel \textit{mention} some politicians in her posts, the politician's \textit{follow}ers might come to interact with her. One might \textit{mention} or \textit{reply} to debate, but \textit{like} always stands for agreement. The relations could reflect some opinions that a user would never tell you verbally. Words could be easily disguised, and there is always a problem called ``the silent majority'', for most people are unwilling to express.

Yet there are some uniqueness of Twitter dataset, bringing about many challenges. It is especially the case when existing approaches are mostly dealing with smaller datasets with much sparser links than ours, such as academic graphs, text-word graphs, and knowledge-graphs. First, our Twitter dataset is large and the links are relatively dense (Section \ref{sec:model}). Some models such as GraphSAGE \cite{hamilton2017inductive} will be super slow sampling our graph. 
Second, labels are extremely sparse, less than $1\%$. Most approaches will suffer from severe over-fitting, and the lack of reliable evaluation. Third, features are always incomplete, for in real-life datasets like Twitter, many accounts are removed or blocked. Fourth, modeling the heterogeneity is nontrivial. Many existing methods designed for homogeneous networks tend to ignore the information brought by the types of links. 

Existing works can not address the above challenges well.
Even though some realized the importance of links \cite{gu2016ideology,conover2011political}, they failed to provide an embedding. Most people learn an embedding by separating the heterogeneous graph into different homogeneous views entirely, and combine them in the very end.

We propose to solve the above-listed problems by \textbf{TIMME} (Twitter Ideology-detection via Multi-task Multi-relational Embedding), a model good at handling sparsely-labeled large graph, utilizing multiple relation types, and optionally dealing with missing features. 
Our code with data is released on Github at \url{https://github.com/PatriciaXiao/TIMME}. 
Our major contributions are: 
\begin{itemize}[leftmargin=*]
\item We propose \textbf{TIMME} for ideology detection on Twitter, whose encoder captures the interactions between different relations, and decoder treats different relations separately while measuring the importance of each relation to ideology detection.
\item The experimental results have proved that \textbf{TIMME} outperforms the state-of-the-art models. Case studies showed that conservative voice is typically under-represented on Twitter. There are also many findings on the relations' interactions.
\item The large-scale dataset we crawled, cleaned, and labeled (Appendix \ref{append_sec:dataprep_details}) provides a new benchmark to study heterogeneous information networks.
\end{itemize}

In this paper, we will walk through the related work in Section \ref{sec:ref}, introduce the preliminaries and the definition of the problem we are working on in Section \ref{sec:problem}, followed by the details of the model we propose in Section \ref{sec:model}, experimental results and discussions in Section \ref{sec:experiment}, and Section \ref{sec:conclusion} for conclusion.

\section{related work}\label{sec:ref}

\subsection{Ideology Detection}

Ideology detection in general could be naturally divided into two directions, based on the targets to predict: of the politicians \cite{poole1985spatial,nguyen2015tea,clinton2004statistical}, and of the ordinary citizens \cite{achen1975mass,kuhn2019national,baly2019multi,gu2016ideology,iyyer2014political,kannangara2018mining,preoctiuc2017beyond,martini2019trust,chen2017opinion,johnson2016identifying,conover2011predicting}. The work conducted on ordinary citizens could also be categorized into two types according to the source of data being used: intentionally collected via strategies like survey \cite{achen1975mass,kuhn2019national}, and directly collected such as from news articles \cite{baly2019multi} or from social networks \cite{gu2016ideology,iyyer2014political,kannangara2018mining}. Some studies take advantages from both sides, asking self-reported responses from a group of users selected from social networks \cite{preoctiuc2017beyond}, and some researchers admitted the limitations of survey experiments \cite{martini2019trust}. 
Emerging from social science, probabilistic models have been widely used for such kinds of analysis since the early $1980$s~\cite{poole1985spatial,gu2016ideology,baly2019multi}. On the other hand, on social network datasets, it is quite intuitive trying to extract information from text data to do ideology-detection \cite{iyyer2014political,kannangara2018mining,chen2017opinion,johnson2016identifying,conover2011predicting}, only a few paid attention to links \cite{gu2016ideology,conover2011political}. Our work differs from them all, since: (1) unlike probabilistic models, we use GNN approaches to solve this problem, so that we take advantage of the high-efficient computational resources, and we have the embeddings for further analysis; (2) we focus on \textbf{relations} among users, and proved how telling those relations are.

\subsection{Graph Neural Networks (GNN)}

\subsubsection{Graph Convolutional Networks (GCN)}\label{subsubsec:gcn}
Inspired by the great success of convolutional neural networks (CNN), researchers have been seeking for its extension onto information networks \cite{defferrard2016convolutional,kipf2016semi} to learn the entities' embeddings.
The Graph Convolutional Networks (GCN) \cite{kipf2016semi} could be regarded as an approximation of spectral-domain convolution of the graph signals. 
A deeper insight~\cite{li2018deeper} shows that the key reason why GCN works so well on classification tasks is that its operation is a form of Laplacian smoothing, and concludes the potential over-smoothing problem, as well as emphasizes the harm of the lack of labels.

GCN convolutional operation could also be viewed as sampling and aggregating of the neighborhood information, such as GraphSAGE \cite{hamilton2017inductive} and FastGCN \cite{chen2018fastgcn}, enabling training in batches. 
To improve GraphSAGE's expressiveness, GIN \cite{xu2018powerful} is developed, enabling more complex forms of aggregation.
In practice, due to the sampling time cost brought by our links' high density, GIN, GraphSAGE and its extension onto heterogeneous information network such as HetGNN \cite{zhang2019heterogeneous} and GATNE \cite{cen2019representation} are not very suitable on our datasets.

The relational-GCN (r-GCN) \cite{schlichtkrull2018modeling} extends GCN onto heterogeneous information networks. A very large number of relation-types $|\mathcal{R}|$ ends up in overwhelming parameters, thus they put some constraints on the weight matrices, referred to as weight-matrix decomposition.
GEM \cite{liu2018heterogeneous} is almost a special case of r-GCN. Unfortunately, their code is kept confidential. According to the descriptions in their paper, they have a component of similar use as the attention weights $\alpha$ in our encoder, but it is treated as a free parameter.

Another way of dealing with multiple link types is well-represented by SHINE \cite{wang2018shine}, who treats the heterogeneous types of links as separated homogeneous links, and combines embeddings from all relations in the end. SHINE did not make good use of the multiple relations to its full potential, modeling the relations without allowing complex interactions among them.
GTN \cite{yun2019graph} is similar with SHINE in splitting the graph into separate views and combining the output at the very end. Besides, 
GTN uses meta-path, thus is potentially more expressive than SHINE, but would rely heavily on the quality and quantity of the meta-paths being used.

\subsubsection{Graph Attention Networks}
Graph Attention Networks (GAT) \cite{velivckovic2017graph} is another nontrivial direction to go under the topic of graph neural networks. It incorporates attention into propagation by applying self-attention on the neighbors. Multi-head mechanism is often used to ensure stability.

An extension of GAT on heterogeneous information networks is Heterogeneous Graph Attention Network, HAN~\cite{wang2019heterogeneous}. Beside inheriting the node-level attention from GAT, it considers different relation types by sampling its neighbors from different meta-paths. It first conducts type-specific transformation and compute the importance of neighbors of each node. After that, it aggregates the coefficients of all neighbor nodes to update the current node's representation. In addition, to obtain more comprehensive information, it conducts semantic-level attention, which takes the result of node-level attention as input and computes the importance of each meta-path.
We use HAN as an important baseline in our experiments.

\subsection{Multi-Task Learning (MTL)}\label{subsec:mtl}

In multi-task learning (MTL) settings, there are multiple tasks sharing the same inductive bias jointly trained. Ideally, the performance of every task should benefit from leveraging auxiliary knowledge from each other.
As is concluded in an overview \cite{ruder2017overview}, MTL could be applied with or without neural network structure. On neural network structure, the most common approach is to do hard parameter-sharing, where the tasks share some hidden layers. The most common way of optimizing an MTL problem is to solve it by joint-training fashion, with joint loss computed as a weighted combination of losses from different tasks \cite{kendall2018multi}. It has a very wide range of applications, such as the DMT-Demographic Models \cite{vijayaraghavan2017twitter} where multiple aspects of Twitter data (e.g. text, images) are fed into different tasks and trained jointly. Aron and Nirmal et al. \cite{culotta2015predicting} also apply MTL on Twitter, separating the tasks by user categories. Our multi-task design differs from theirs, and treat node classification and link prediction on different relation types as different tasks.
\section{Problem Definition}\label{sec:problem}

Our goal is to predict Twitter users' ideologies, by learning the ideology embedding of users in a political-centered social network. 

\begin{definition}{\textbf{(Heterogeneous Information Network)}}
Following previous work \cite{sun2012mining}, we say that an information network $\mathcal{G} = \{ \mathcal{V}, \mathcal{E} \}$, where number of vertices is $|\mathcal{V}| = N$, is a \textbf{heterogeneous information network}, when there are $|\mathcal{T}| = T$ types of vertices, $|\mathcal{R}| = R$ types of edges, and $\max(T, R) > 1$. $\mathcal{G}$ could be represented as $\mathcal{G} = \{ \{\mathcal{V}_1, \mathcal{V}_2, \dots \mathcal{V}_T \}, \{ \mathcal{E}_1, \mathcal{E}_2, \dots, \mathcal{E}_R \} \}$
\end{definition}

Each possible edge from the $i^{th}$ node to the $j^{th}$, represented as $e_{ij} \in \mathcal{E}$ has a weight value $w_{ij} > 0$ associated to it, where $w_{ij} = 0$ representing $e_{ij} \notin \mathcal{E}$. In our case, $\mathcal{G}$ is a directed graph. In general, we have $\langle v_i, v_j \rangle \not\equiv \langle v_j, v_i \rangle$ and $w_{ij} \not\equiv w_{ji}$.

Twitter data $\mathcal{G}_{Twitter}$ contains $T = 1$ type of entities (users), and $R = 5$ different types of edges (relations) among the entities, namely \textit{follow}, \textit{retweet}, \textit{like}, \textit{mention}, \textit{reply}.
\begin{displaymath}
\mathcal{G}_{Twitter} = \{ \mathcal{V}, \{ \mathcal{E}_1, \mathcal{E}_2, \mathcal{E}_3, \mathcal{E}_4, \mathcal{E}_5 \} \}
\end{displaymath}

Detailed description about Twitter data is included in Appendix \ref{append_sec:dataprep_details}, and we call the subgraph we selected from Twitter-network a \textit{political-centered social network}, which is defined as follows:

\begin{definition}{\textbf{(Political-Centered Social Network)}}
The political-centered social network is a special case of directed heterogeneous information network.
With a pre-defined politicians set $\mathcal{P}$, in our selected heterogeneous network $\mathcal{G}_{Twitter}$, $\forall e = \langle v_i, v_j \rangle \in \mathcal{E}_r$ where $r \in \{ 1, 2, \dots, R \}$, there has to be either $v_i \in \mathcal{P}$ or $v_j \in \mathcal{P}$. All the politicians in this dataset have ground-truth labels indicating their political stance. The political-centered social networks are represented as $\mathcal{G}_{p}$.
\end{definition}

We would like to leverage the information we have to learn the representation of the users, which could help us reveal their ideologies. Due to the lack of Independent representatives (only two in total), we consider the binary-set labels only: \{ \textit{liberal}, \textit{conservative} \}. \textit{Democratic} on liberal side, \textit{Republican} for conservative.

\begin{definition}{\textbf{(Multi-task Multi-relational Network Embedding)}}
Given a network $\mathcal{G}_p = \{ \mathcal{V}, \{ \mathcal{E}_1, \mathcal{E}_2, \mathcal{E}_3, \mathcal{E}_4, \mathcal{E}_5 \} \}$ where the number of nodes is $|\mathcal{V}| = N$, the goal of TIMME is to learn such a representation $h_i \in \mathbb{R}^d$ where $d \ll N$ for $\forall v_i \in \mathcal{V}$, that captures the categorical information of nodes, such as their ideology tendencies. As a measurement, we want the representation $H \in \mathbb{R}^{N \times d}$, to success on both node-classification and link-prediction.
\end{definition}

\section{Methodology}\label{sec:model}

\begin{figure}
\centering
\includegraphics[width=0.85\linewidth]{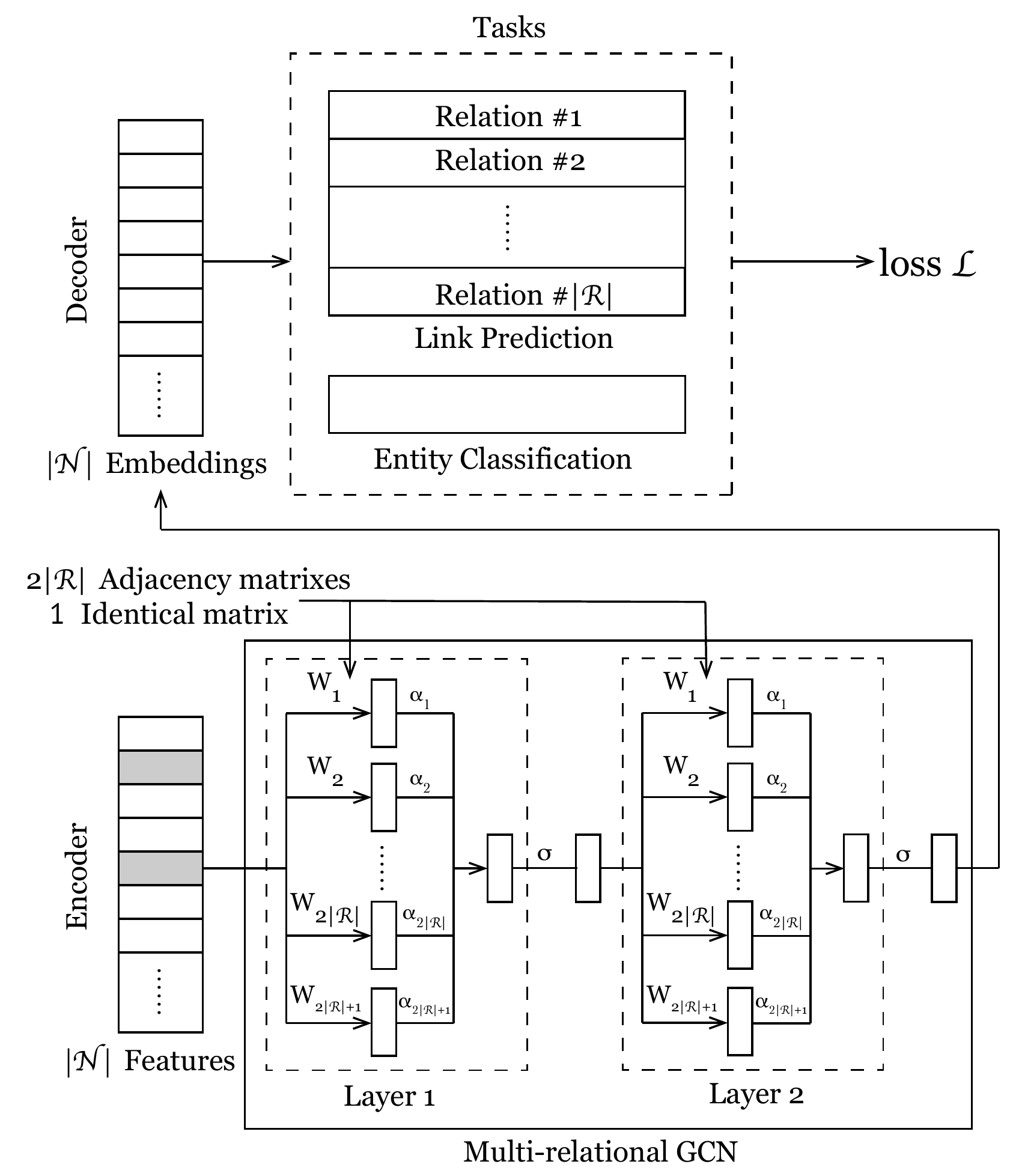}
\caption{The general architecture of our model, with the encoder shown in details. Grey blocks represent missing features. Our model can either handle them by treating them as learnable parameters, or use one-hot features.}
\label{fig:general_architecture}
\end{figure}
The general architecture of our proposed model is illustrated in Figure~\ref{fig:general_architecture}. It contains two components: encoder and decoder. The encoder contains two multi-relational convolutional layers. The output of the encoder is passed on to the decoder, who handles the downstream tasks.

\subsection{Multi-Relation Encoder}\label{subsec:encoder}

As mentioned before in Section \ref{sec:intro}, the challenges faced by the encoder part are the large data scale, the heterogeneous link types, and the missing features.

GCN is very effective in learning the nodes' embeddings, especially good at classification tasks. Meanwhile, it is also naturally efficient, in terms of handling the large amount of vertices $N$.

Random-walk-based approaches such as node2vec~\cite{grover2016node2vec} with time complexity $\mathcal{O}(a^2N)$, where $a$
is the average degree of the graph, suffer from the relatively-high degree in our dataset.
On the other hand, GCN-based approaches are naturally efficient here. Like is analyzed in Cluster-GCN \cite{chiang2019cluster}, the time complexity of the standard GCN model is $\mathcal{O}(L\lVert A \rVert_0 F + L N F^2)$, where $L$ is the number of layers, $\lVert A \rVert_0$ the number of non-zeros in the adjacency matrix, $F$ the number of features. Note that the time complexity increases linearly when $N$ increases.
\begin{figure*}
\centering
\includegraphics[width=1.0\linewidth]{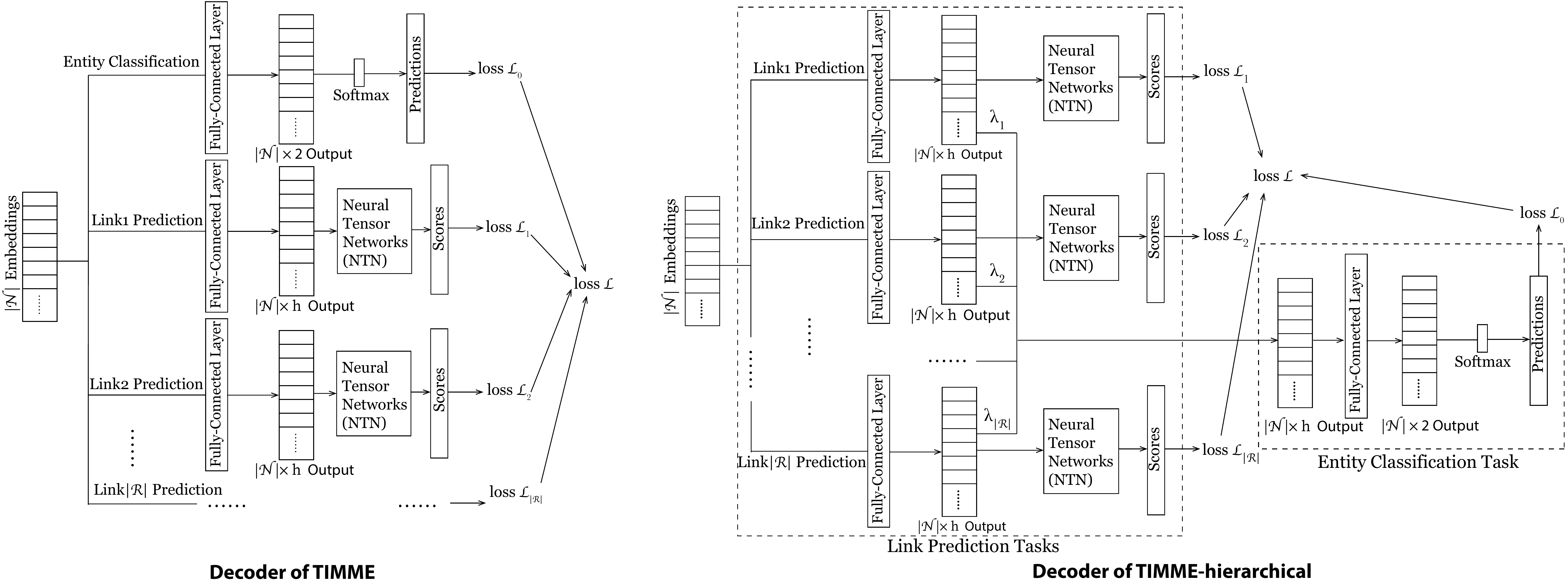}
\caption{The two types of decoder in our multi-task framework, referred to as \textit{TIMME} and \textit{TIMME-hierarchical}. }
\label{fig:decoder}
\end{figure*}
A GCN model's layer-wise propagation could be written as:
\begin{displaymath}
H^{(l+1)} = \sigma\Big( \hat{A} H^{(l)} W^{(l)} \Big)
\end{displaymath}
$\hat{A} = \tilde{D}^{\frac{1}{2}} (A + I_N) \tilde{D}^{\frac{1}{2}}$, where $\tilde{D}$ is defined as the diagonal matrix and $A$ the adjacency matrix. $D_{ii}$, the diagonal element $d_i$, is equal to the sum of all the edges attached to $v_i$; $H^{(l)} \in \mathbb{R}^{N \times d^{(l)}}$ is the $d^{(l)}$-dimensional representation of the $N$ nodes at the $l^{th}$ layer; $W^{(l)} \in \mathbb{R}^{d^{(l)} \times d^{(l+1)}}$ is the weight parameters at layer $l$ which is similar with that of an ordinary MLP model \footnote{\href{https://en.wikipedia.org/wiki/Multilayer_perceptron}{MLP here refers to Multi-layer Perceptron.}}. In a certain way, $\hat{A}$ could be viewed as $A$ after being normalized.

We propose to model the heterogeneous types of links and their interactions in the encoder. Otherwise, if we split the views like many others did, the model will never be expressive enough to capture the interactions among relations.
For any given political-centered graph $\mathcal{G}_P$, let's denote the total number of nodes $|\mathcal{V}| = N$, the number of relations $|\mathcal{R}| = R$, the set of nodes $\mathcal{V}$, the set of relations $\mathcal{R}$, and $\mathcal{E}_r$ being the set of links under relation $r \in \mathcal{R}$. Representation being learned after layer $l$ ($l \in \{ 1, 2\}$) is represented as $H^{(l)} \in \mathbb{R}^{N \times d^{(l)}}$, and the input features form the matrix $H^{(0)} \in \mathbb{R}^{N \times d^{(0)}}$. $\hat{\mathcal{R}}$ where $|\hat{\mathcal{R}}| = 2R+1$ represents all relations in the original direction ($R$), the relations in reversed direction ($R$), and an identical-matrix relation ($1$).
Our dataset has $|\mathcal{R}| = R = 5$, so it should be fine not to conduct a weight-matrix decomposition like r-GCN \cite{schlichtkrull2018modeling}. We model the layer-wise propagation at Layer $l+1$ as: 
\begin{displaymath}
H^{(l+1)} = \sigma\Big( \sum_{r \in \hat{\mathcal{R}}} \alpha_r \hat{A}_r H^{(l)} W_r^{(l)} \Big)
\end{displaymath}
where $H^{(l)} \in \mathbb{R}^{N \times d^{(l)}}$ is used to denote the representation of the nodes after the $l^{th}$ encoder layer, 
and the initial input feature is $H^{(0)}$. $\hat{A}_r = \tilde{D}_r^{\frac{1}{2}} (A_r + I_N) \tilde{D}_r^{\frac{1}{2}} $ is defined in similar way as $\hat{A}$ in GCN, but it is calculated per relation.
The activation function $\sigma$ we use is ReLU. By default, $\alpha = = [\alpha_1, \dots \alpha_r \dots]^T \in \mathbb{R}^{2R+1}$ is calculated by scaled dot-product self-attention over the outputs of $H^{(l+1)}_r = \hat{A}_r H^{(l)} W_r^{(l)}$:
\begin{displaymath}
A = Attention(Q, K, V) = softmax\big(\frac{QK^T}{\sqrt{d}}\big)V \in \mathbb{R}^{ (2R+1) \times d}
\end{displaymath}
where $Q = K = V \in \mathbb{R}^{(2R+1) \times d}$ comes from the $2R+1$ matrices $H^{(l+1)}_r \in \mathbb{R}^{N \times d}$, stacking up as $O \in \mathbb{R}^{(2R+1) \times N \times d}$, taking an average over the $N$ entities. 
We calculate an attention to apply to the $2R+1$ outputs as:
\begin{displaymath}
\alpha = softmax\ \text{sum}_{\text{col}}\Big(\frac{QK^T}{\sqrt{d}}\Big) \in \mathbb{R}^{2R+1}
\end{displaymath}
where $\text{sum}_{\text{col}}(X)$ takes the sum of each column in $X \in \mathbb{R}^{d_1\times d_2}$ and ends up in a vector $ \in \mathbb{R}^{d_2}$.

The last problem to solve is that the initial features $H^{(0)}$ is often incomplete in real life. In most cases, people would go by one-hot features or randomized features. But we want to enable our model to use the real features, even if the real-features are incomplete.
Inspired by graph representation learning strategies such as LINE \cite{tang2015line}, we proposed to treat the unknown features as trainable parameters. That is, for a graph $\mathcal{G}_p$ whose vertice set is $\mathcal{V}$, $\mathcal{V}_{featured} \bigcap \mathcal{V}_{featureless} = \varnothing$ and $\mathcal{V}_{featured} \bigcup \mathcal{V}_{featureless} = \mathcal{V}$, for any node with valid feature $\forall v_i \in \mathcal{V}_{featured}$, the node's feature vector $H_i^{(0)}$ is known and fixed. For $\forall v_j \in \mathcal{V}_{featureless}$, the corresponding row vector $H_j^{(0)}$ is unknown and treated as a trainable parameter. The generation of the features will be discussed in the Appendix \ref{append_sec:dataprep_details}. In brief, TIMME can handle any missing input feature.

\subsection{Multi-Task Decoder}

We propose \textbf{TIMME} as a multi-task learning model such that the sparsity of the labels could be overcome with the help of the link information. As is shown in Figure \ref{fig:decoder}, we propose two architectures of the multi-task decoder. When we test it on a  single-task $i$, we simply disable the remaining losses but a single $\mathcal{L}_i$, and name our model in single-task mode \textbf{TIMME-single}.

$\mathcal{L}_0$ is defined the same way as was proposed in \cite{kipf2016semi}, in our case a binary cross-entropy loss:
\begin{displaymath}
\mathcal{L}_0 = -\sum_{y \in Y_{train}}\big(y\log(y)+(1-y)\log(1-y)\big)
\end{displaymath}
where $Y_{train}$ contains the labels in the training set we have. 

$\mathcal{L}_1, \dots \mathcal{L}_R$ are link-prediction losses, calculated by binary cross-entropy loss between link-labels and the predicted link scores' logits.
To keep the links asymmetric, we used Neural Tensor Network (NTN) structure \cite{socher2013reasoning}, with simplification inspired by DistMult \cite{yang2014embedding}.
We set the number of slices be $k = 1$ for $W_r \in \mathbb{R}^{d \times d \times k}$, omitting the linear transformer $U$, and restricting the weight matrices $W_r$ each being a diagonal matrix. For convenience, we refer to this link-prediction cell as \textbf{TIMME-NTN}. Consider triplet $(v_i, r, v_j)$, and denote the encoder output of $v_i, v_j \in \mathcal{V}$ as $h_i, h_j \in \mathbb{R}^d$, the score function of the link is calculated as:
\begin{displaymath}
s(i,r,j) = h_i W_r h_j + V\begin{bmatrix} h_i \\ h_j \end{bmatrix} + b
\end{displaymath}
where $W_r \in \mathbb{R}^{d \times d}$ is a diagonal matrix for any $\forall r \in \mathcal{R}$. $W_r$, $V \in \mathbb{R}^{2d}$ and $b \in \mathbb{R}$ are all parameters to be learned. Group-truth label of a positive (existing) link is $1$, otherwise $0$.

The first decoder-architecture \textbf{TIMME} sums all $R+1$ losses as $\mathcal{L} = \sum_{i = 0}^R \mathcal{L}_i$. Without average, each task's loss is directly proportional to the amount of data points sampled at the current batch. Low-resource tasks will take a smaller portion. This is the most straightforward design of a MTL decoder. 

The second, \textbf{TIMME-hierarchical}, has $\lambda = [\lambda_1, \dots, \lambda_{|\mathcal{R}|}]^T$ being computed via self-attention on the average embedding over the $R$ link-prediction task-specific embeddings.
Here, $\mathcal{L} = \sum_{i = 0}^R \mathcal{L}_i$ is the same with \textbf{TIMME}. \textbf{TIMME-hierarchical} essentially derives the node-label information from the link relations, thus provides some insights on each relation's importance to ideology prediction. \textbf{TIMME}, \textbf{TIMME-hierarchical}, \textbf{TIMME-single} models share \textit{exactly the same} encoder architecture.

\section{Experiments}\label{sec:experiment}

In this section, we introduce the dataset we crawled, cleaned and labeled, together with our experimental results and analysis.

\subsection{Data Preparation}\label{subsec:dataprep}

\subsubsection{Data Crawling}
\begin{table}
\centering
\begin{tabular}{lcccc} 
\toprule 
 & PureP & P50 & P20$\sim$50 & P+all \\
\midrule 
\#\ User & 583 & 5,435 & 12,103 & 20,811 \\
\#\ Link & 122,347 & 1,593,721 & 1,976,985 & 6,496,107 \\
\#\ Labeled User & 581 & 759 & 961 & 1,206 \\
\#\ Featured User & 579 & 5,149 & 11,725 & 19,418 \\
\midrule 
\#\ Follow-Link & 59,073 & 529,448 & 158,746 & 915,438 \\
\#\ Reply-Link & 1,451 & 96,757 & 121,133 & 530,598 \\
\#\ Retweet-Link & 19,760 & 311,359 & 595,030 &  1,684,023 \\
\#\ Like-Link & 14,381 & 302,571 & 562,496 & 1,794,111 \\
\#\ Mention-Link & 27,682 & 353,586 & 539,580 & 1,571,937 \\
\bottomrule 
\end{tabular}
\caption{Descriptive statistics of the three selected subsets of our dataset.}
\label{tab::stat}
\vspace{-6mm}
\end{table}

The statics of the political-centered social network datasets we have are listed in Table \ref{tab::stat}. Data prepared is described in Appendix \ref{append_sec:dataprep_details}, ready by April, 2019. In brief, we did:
\begin{enumerate}[leftmargin=*]
    \item Collecting some Twitter accounts of the politicians $\mathcal{P}$;
    \item For every politician $\forall p \in \mathcal{P}$, crawl her/his most-recent $s$ followers and $s$ followees, putting them in a candidate set $\mathcal{C}$.
    \item For every candidate $c \in \mathcal{C}$, we also crawl their most-recent $s$ followers to make the \textit{follow} relation more complete.
    \item For every user $u \in \mathcal{P} \cup \mathcal{C}$, crawl their tweets as much as possible, until we hit the limit ($\approx 3,200$) set by Twitter API.
    \item From the followers \& followees we collect \textit{follow} relation, from the tweets we extract: \textit{retweet}, \textit{mention}, \textit{reply}, \textit{like}.
    \item Select different groups of users from $\mathcal{C}$, based on how many connections they have with members in $\mathcal{P}$, and making those groups into the $4$ subsets, as is shown in Table \ref{tab::stat}.
    \item We filter the relations within any selected group so that if a relation $e = \langle v_i, v_j \rangle \in \mathcal{G}_p$, there must be $v_i \in \mathcal{G}_p$ and $v_j \in \mathcal{G}_p$.
\end{enumerate}
Our four datasets represent different user groups. \textbf{PureP} contains only the politicians. \textbf{P50} contains politicians and users keen on political affairs. \textbf{P20$\sim$50} is politicians with the group of users who are of moderate interests on politics. \textbf{P+all} is a union set of the three, plus some randomly-selected outliers of politics. \textbf{P+all} is the most challenging subset to all models. More details on the dataset, including how we generated features and how we tried to get more labels, are all described in details in Appendix \ref{append_sec:dataprep_details}.

\subsection{Performance Evaluation}

In practice, we found that we do not need any features for nodes, and use one-hot encoding vector as initial feature.

We split the train, validation, and test set of node labels by 8:1:1, keep it the same across all datasets and throughout all models, measuring the labels' prediction quality by F1-score and accuracy. For link-prediction tasks, we split all positive links into training, validation, and testing sets by 85:5:10, keeping same portion across all datasets and all models, evaluating by ROC-AUC and PR-AUC. \footnote{AUC refers to Area Under Curve, PR for precision-recall curve, ROC for receiver operating characteristic curve.} 

\subsubsection{Baseline Methods}

{
\scriptsize
\begin{table}
\centering
\begin{tabular}{lcccc} 
\toprule 
Model & PureP & P50 & P20$\sim$50 & P+all \\
\midrule 
GCN & \textbf{1.0000}/\textbf{1.0000} & 0.9600/0.9600 & 0.9895/0.9895 & 0.9076/0.9083\\
r-GCN & \textbf{1.0000}/\textbf{1.0000} & 0.9733/0.9733 & 0.9895/0.9895 & 0.9327/0.9333 \\
HAN & 0.9825/0.9824 & 0.9466/0.9467 & 0.9789/0.9789 & 0.9238/0.9250\\
\midrule 
TIMME-single & \textbf{1.0000}/\textbf{1.0000} & 0.9733/0.9733 & 0.9895/0.9895 & 0.9333/0.9324 \\
TIMME & 0.9825/0.9824 & \textbf{0.9867}/\textbf{0.9867} & \textbf{1.0000}/\textbf{1.0000} & 0.9495/0.9500 \\
TIMME-hierarchical & \textbf{1.0000}/\textbf{1.0000} & 0.9733/0.9780 & 0.9895/0.9895 & \textbf{0.9580}/\textbf{0.9583} \\
\bottomrule 
\end{tabular}
\caption{Node classification measured by F1-score/accuracy.}
\label{tab::classification}
\vspace{-3mm}
\end{table}
}

{
\scriptsize
\begin{table}
\centering
\begin{tabular}{llccc} 
\toprule 
Model & PureP & P50 & P20$\sim$50 & P+all \\
\midrule
\multicolumn{5}{c}{Follow Relation} \\
\midrule 
GCN+ & 0.8696/0.6167 & 0.9593/0.8308 & 0.9870/0.9576  & 0.9855/0.9329 \\
r-GCN & 0.8596/0.6091 & 0.9488/0.8023 & 0.9872/0.9537 & 0.9685/0.9201\\
HAN+ & \textbf{0.8891}/\textbf{0.7267} & 0.9598/0.8642 & 0.9620/0.8850 & 0.9723/0.9256 \\
\midrule 
TIMME-single & 0.8809/0.6325 & 0.9717/0.8792 & 0.9920/0.9709 & 0.9936/0.9696 \\
TIMME & 0.8763/0.6324 & \textbf{0.9811}/\textbf{0.9154} & 0.9945/0.9799 & 0.9943/0.9736 \\
TIMME-hierarchical & 0.8812/0.6409 & 0.9809/0.9145 & \textbf{0.9984}/\textbf{0.9813} & \textbf{0.9944}/\textbf{0.9739}  \\
\midrule
\multicolumn{5}{c}{Reply Relation} \\
\midrule 
GCN+ & 0.8602/0.7306 & 0.9625/0.9022 & 0.9381/0.8665 & 0.9705/0.9154 \\
r-GCN & 0.7962/0.6279 & 0.9421/0.8714 & 0.8868/0.7815 & 0.9640/0.9085 \\
HAN+ & 0.8445/0.6359 & 0.9598/0.8616 & 0.9495/0.8664 & 0.9757/0.9210\\
\midrule 
TIMME-single & 0.8685/0.7018 & 0.9695/0.9307 & 0.9593/0.9070 & 0.9775/0.9508 \\
TIMME & 0.9077/0.8004 & \textbf{0.9781}/\textbf{0.9417} & \textbf{0.9747}/\textbf{0.9347} & 0.9849/0.9612  \\
TIMME-hierarchical & \textbf{0.9224}/\textbf{0.8152} & 0.9766/0.9409 & 0.9737/0.9341 & \textbf{0.9854}/\textbf{0.9629} \\
\midrule
\multicolumn{5}{c}{Retweet Relation} \\
\midrule 
GCN+ & 0.8955/0.7145 & 0.9574/0.8493 & 0.9351/0.8408 & 0.9724/0.9303 \\
r-GCN & 0.8865/0.6895 & 0.9411/0.8084 & 0.9063/0.7728 & 0.9735/0.9326 \\
HAN+ & 0.7646/0.6139 & 0.9658/0.9213 & 0.9478/0.8962 & 0.9750/0.9424\\
\midrule 
TIMME-single & 0.9015/ 0.7202 & 0.9754/0.9127 & 0.9673/0.9073 & 0.9824/0.9424 \\
TIMME & 0.9094/0.7285 & 0.9779/0.9181 & \textbf{0.9772}/\textbf{0.9291} & 0.9858/0.9511 \\
TIMME-hierarchical & \textbf{0.9105}/\textbf{0.7344} & \textbf{0.9780}/\textbf{0.9190} & 0.9766/0.9275 & \textbf{0.9869}/\textbf{0.9543} \\
\midrule
\multicolumn{5}{c}{Like Relation} \\
\midrule 
GCN+ & 0.9007/0.7259 & 0.9527/0.8499 & 0.9349/0.8400 & 0.9690/0.9032 \\
r-GCN &0.8924/0.7161 &  0.9343/0.7966 & 0.9038/0.7681 & 0.9510/0.8945 \\
HAN+ & 0.8606/0.6176 & 0.9733/0.8851 & 0.9611/0.9062 & \textbf{0.9894}/0.9481 \\
\midrule 
TIMME-single & 0.9113/0.7654 & 0.9725/0.9119 & 0.9655/0.9069 & 0.9796/0.9374 \\
TIMME & 0.9249/0.7926 & \textbf{0.9753}/0.9171 & \textbf{0.9759}/\textbf{0.9292} & 0.9846/0.9504 \\
TIMME-hierarchical & \textbf{0.9278}/\textbf{0.7945} & 0.9752/\textbf{0.9175} & 0.9752/0.9271 & 0.9851/\textbf{0.9518} \\
\midrule
\multicolumn{5}{c}{Mention Relation} \\
\midrule 
GCN+ & 0.8480/0.6233 & 0.9602/0.8617 & 0.9261/0.8170 & 0.9665/0.8910 \\
r-GCN &  0.8312/0.6023& 0.9382/0.7963 & 0.8938/0.7563 & 0.9640/0.8902\\
HAN+ & \textbf{0.9000}/\textbf{0.7206} & 0.9573/0.8616 & 0.9574/0.8891 & 0.9724/0.9119 \\
\midrule 
TIMME-single & 0.8587/0.6502 & 0.9713/0.8981 & 0.9614/0.8923 & 0.9725/0.9096 \\
TIMME & 0.8684/0.6689 & 0.9730/0.9035 & \textbf{0.9730}/\textbf{0.9185} & 0.9839/0.9446 \\
TIMME-hierarchical & 0.8643/0.6597 & \textbf{0.9732}/\textbf{0.9046} & 0.9723/0.9166 & \textbf{0.9846}/\textbf{0.9463} \\
\bottomrule 
\end{tabular}
\caption{Link-prediction measured by ROC-AUC/PR-AUC.}
\label{tab::link_prediction}
\vspace{-6mm}
\end{table}
}

We have explored a lot of possible baseline models.
Some methods we mentioned in section \ref{sec:ref}, HetGNN \cite{zhang2019heterogeneous}, GATNE \cite{cen2019representation} and GTN \cite{yun2019graph} generally converge $\approx 10 \sim 100$ times slower than our model on any task. GraphSAGE \cite{hamilton2017inductive} is not very suitable on our dataset. Moreover, other well-designed models such as GIN \cite{xu2018powerful} are way too different from our approach at a very fundamental level, thus are not considered as baselines.
Some other methods such as GEM \cite{liu2018heterogeneous} and SHINE \cite{wang2018shine} should be capable of handling the dataset at this scale, but they are not releasing their code to the public, and we can not easily guarantee reproduction.

We decided to use the three baselines: GCN, r-GCN and HAN. They are closely-related to our model, open-sourced, and efficient.
We understand that none of them were specifically designed for social-networks. Early explorations without tuning them resulted in terrible outcomes. To make the comparisons fair, we did a lot of work in hyper-parameter optimization, so that their performances are significantly improved.
The GCN baseline treats all links as the same type and put them into one adjacency matrix.
We also extend the baseline models to new tasks that were not mentioned in their original papers.
We refer to GCN+ and HAN+ as the GCN-base-model or HAN-base-model with \textbf{TIMME-NTN} attached to it. By comparing with GCN/GCN+, we show that reserving heterogeneousity is beneficial. Comparing with r-GCN, we prove that their design is not as suitable for social networks as ours. With HAN/HAN+ we show that, although their model is potentially more expressive, our model still outperforms theirs in most cases, even after we carefully improved it to its highest potential (Appendix \ref{append_sec:baseline}). We did not have to tune the hyper-parameters of \textbf{TIMME} models closely as hard, thanks to its robustness.

HAN+ has an expressive and flexible structure that helps it achieve high in some tasks. The downsides of HAN/HAN+ are also obvious: it easily gets over-fitting, and is extremely sensitive to dataset statistics, with large memory consumption that typically more than \textit{32G} to run tasks on \textbf{P+all}, where TIMME models takes less than \textit{4G} space with the same hidden size and embedding dimensions as the baseline model's settings.

\subsubsection{TIMME}

To stabilize training, we would have to use the step-decay learning rate scheduler, the same with that for ResNet. 
The optimizer we use is Adam, kept consistent with GCN and r-GCN. We do not need input features for nodes, thus our encoder utilizes one-hot embedding by default. 
One of the many advantages of \textbf{TIMME} is how robust it is to the hyper-parameters and all other settings, reflected by that the same default parameter settings serve all experiments well. 
Like many others have done before, to avoid information leakage, whenever we run tasks involving link-prediction, we will remove all link-prediction test-set links from our adjacency matrices.

It is shown in Table \ref{tab::classification} and \ref{tab::link_prediction} that multi-task models \textbf{TIMME} and \textbf{TIMME-hierarchical} are generally better than \textbf{TIMME-single} on most tasks. Even \textbf{TIMME-single} is superior to the baseline models most of the times. TIMME models are stable and scalable. The classification task, despite the many labels we manually added, easily over-estimating the models. Models trained on single node-classification task will easily get over-fitted. If we force them to keep training after convergence, only multi-task \textbf{TIMME models} keep stable. The baselines and \textbf{TIMME-single} suffer from dramatic performance-drop, especially HAN/HAN+. 

\begin{figure}
\centering
\includegraphics[width=1.0\linewidth]{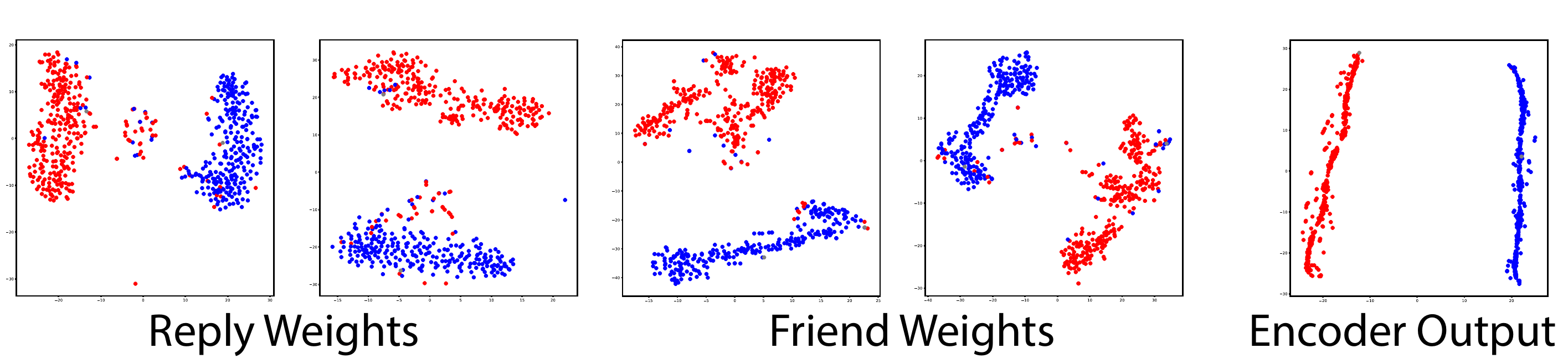}
\caption{t-SNE of matrices onto 2D space. Showing \textit{reply} (and reversed), \textit{friend} (and reversed) weight matrices of the first convolutional layer ($W_r^{(0)}$), and the encoder output embeddings ($H^{(2)}$). Red for ground-truth republican nodes, blue for democratic.}
\label{fig:tsne_weight}
\vspace{-2mm}
\end{figure}

\subsection{Case Studies}

\subsubsection{Selection of Input Features}\label{subsub:feature}

To justify the reason why we do not need any features for nodes, we show the node-classification training-curves of \textbf{TIMME-single} with one-hot features, randomized features, partly-known-partly-randomized features, and with partly-known-partly-trainable features. The results are collected from \textbf{P50} dataset. To make it easier to compare, we have fixed training epochs $300$ for node-classification, and $200$ for follow-relation link-prediction. It is shown that text feature is significantly better than randomized feature, and treating the missing part of the text-generated feature as trainable is better than treat it as fixed randomized feature. However, one-hot feature always outperforms them all, essentially means that relations are more reliable and less noisy than text information in training our network embedding. 
\begin{figure}
\centering
\includegraphics[width=1.0\linewidth]{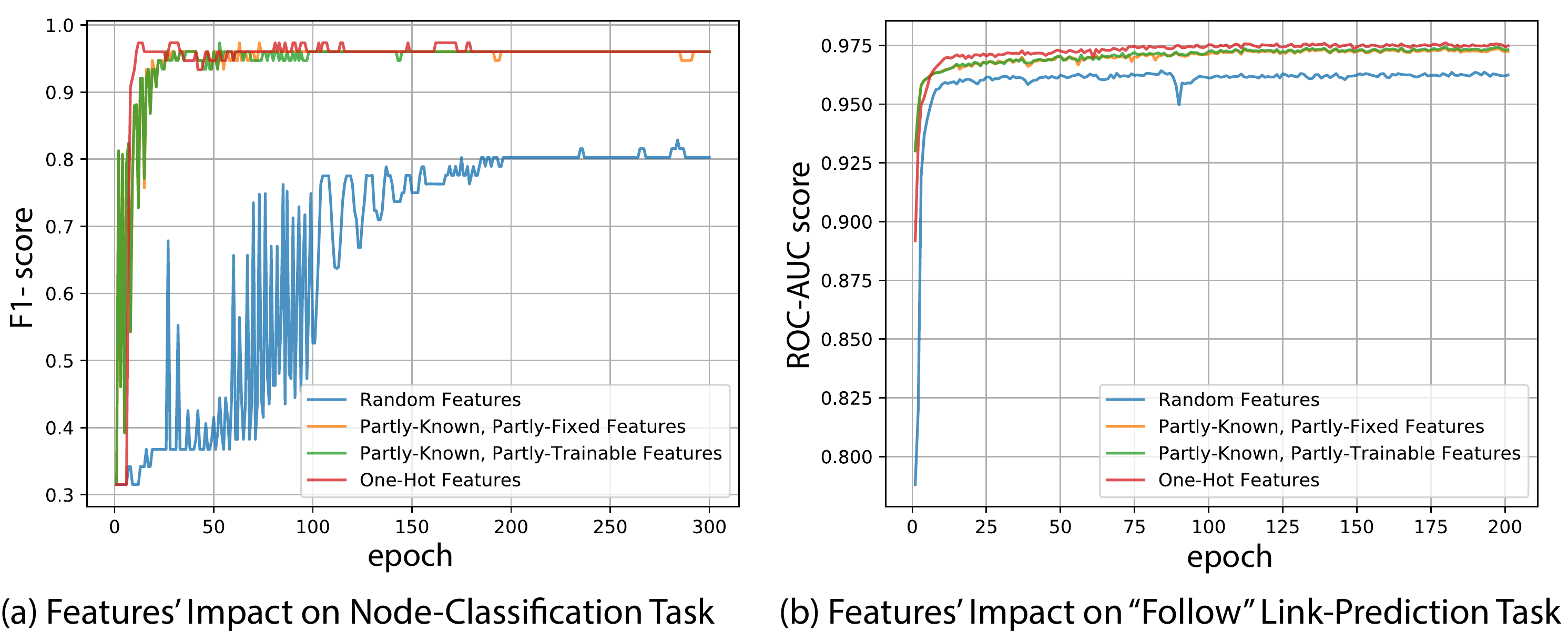}
\caption{Illustration of impact of features. Random features in blue, partly-know partly randomized (and fixed) in yellow, partly-known partly-trainable in green, one-hot in red.}
\label{fig:features}
\vspace{-3mm}
\end{figure}
We have proved in Appendix \ref{append_sec:proof} that the $2R+1$ weight matrices at the first convolutional layer captures the nodes' learned features when using one-hot features. Experimental evidence is shown in Figure \ref{fig:tsne_weight}. 
It shows that although worse than the encoder output, the first embedding layer also captured the features of nodes. 
The embedding comes from epoch $300$, node-classification task on \textbf{PureP}.

\begin{figure}
\centering
\includegraphics[width=1.0\linewidth]{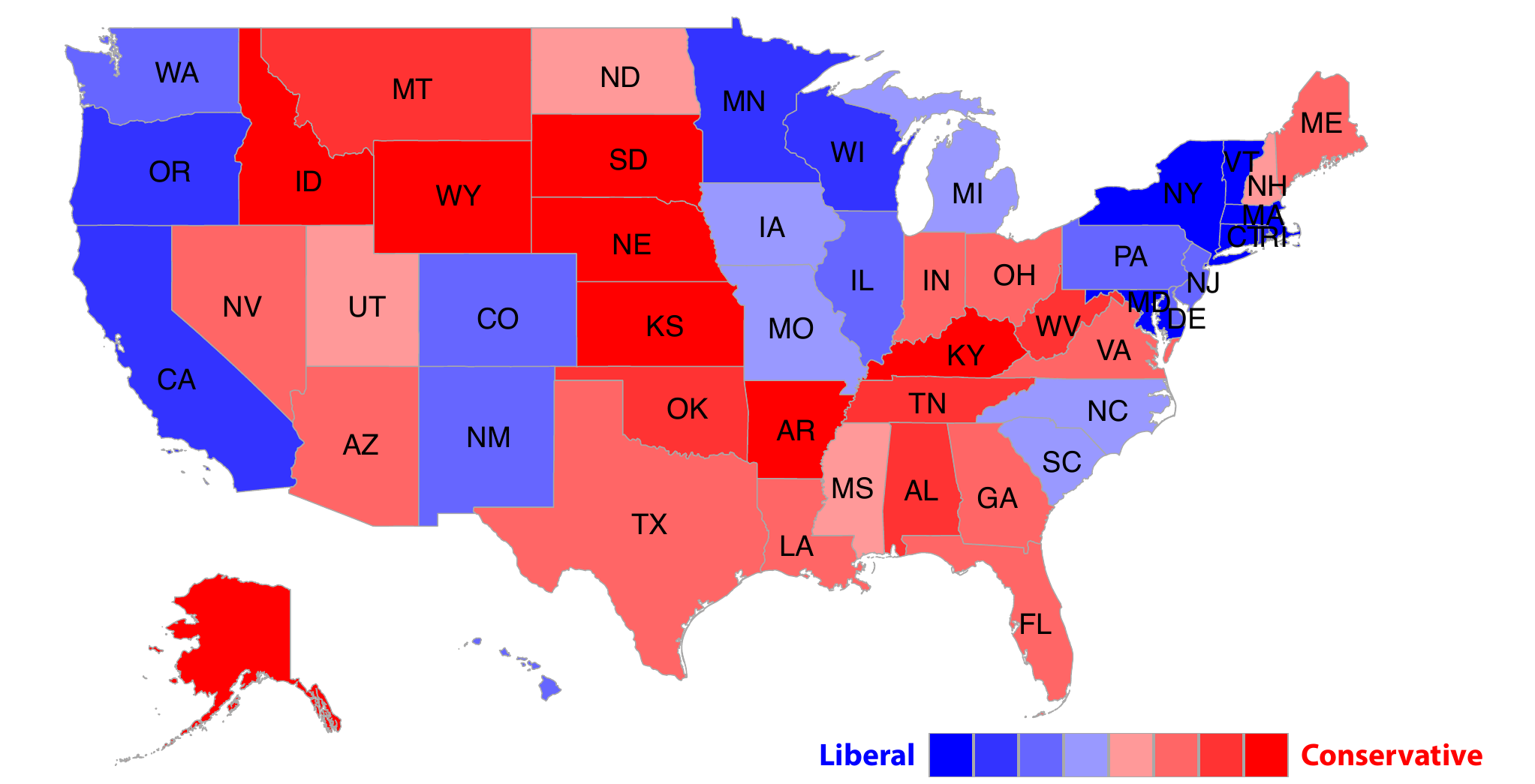}
\caption{Overall ideology on Twitter in each state.}
\label{fig:all_states}
\end{figure}

\begin{figure}
\centering
\includegraphics[width=1.0\linewidth]{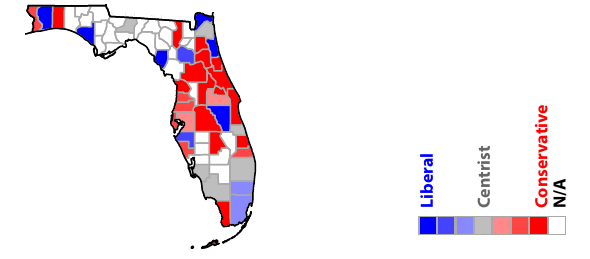}
\caption{Overall ideology on Twitter, Florida (FL).}
\label{fig:florida}
\end{figure}

\begin{figure}
\centering
\includegraphics[width=1.0\linewidth]{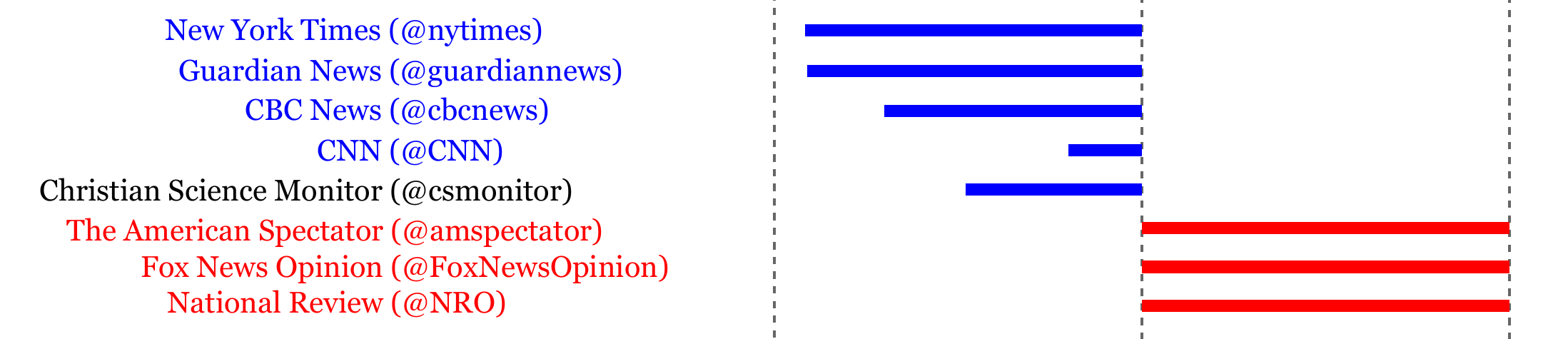}
\caption{The News Agencies' Ideologies. Text colors come from the public's voting online, blue for left and red for right, black for middle (centrist). Length represents the value from the last layer, reflecting the extent.
}
\label{fig:news_agencies}
\vspace{-3mm}
\end{figure}

\begin{figure*}
\centering
\includegraphics[width=1.0\linewidth]{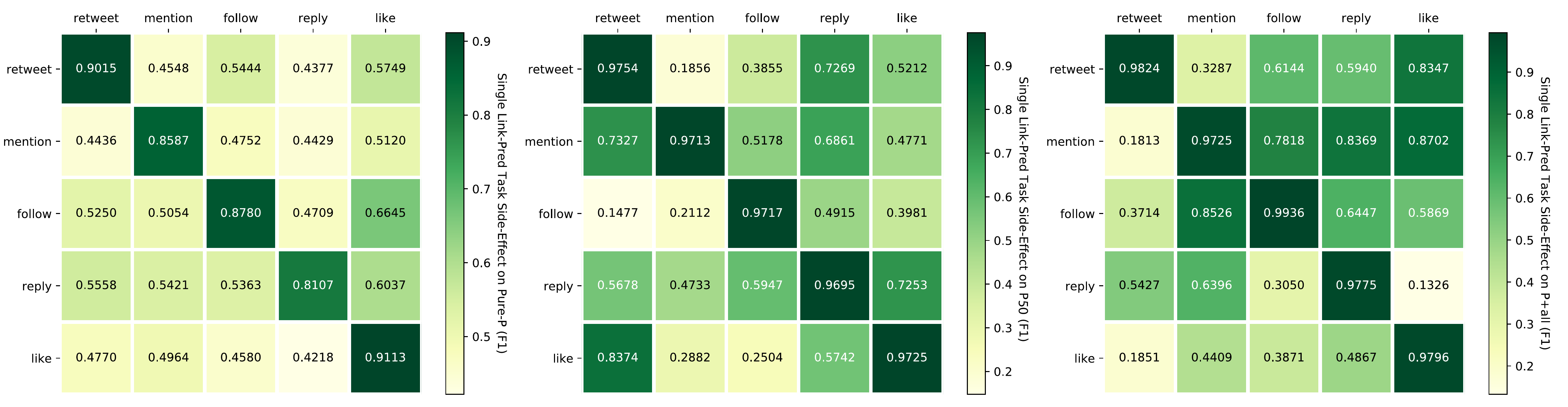}
\caption{The impact of training on single-link-prediction tasks, on Pure-P (left), P50 (middle), P+all (right) dataset respectively. }
\label{fig:relation_relation}
\end{figure*}

\subsubsection{Performance Measurement on News Agency}
A good measurement of our prediction's quality would be on some users with ground-truth tendency, but unlabeled in our dataset. News agents' accounts are typically such users, as is shown in Figure \ref{fig:news_agencies}. Among them we select some of the agencies believed to have clear tendencies.
\footnote{We fetch most of the ground-truth labels of the news agents from the public voting results on \url{https://www.allsides.com/media-bias/media-bias-ratings}, got them \textbf{after} the prediction results are ready.} 
The continuous scores we have for prediction come from the softmax of the last-layer output of our node-classification task, which is in the format of $(prob_{left}, prob_{right})$. Right in the middle represents $(prob_{left}, prob_{right}) = (0.5, 0.5)$, left-most being $(1.0, 0.0)$, right-most $(0.0, 1.0)$.
For most cases, our model's predictions agree with people's common belief. But CNN News is an interesting case. It is believed to be extremely left, but predicted as slightly-left-leaning centrist. Some others have findings supporting our results: CNN is actually only a little bit left-leaning. \footnote{\url{https://libguides.com.edu/c.php?g=649909&p=4556556}} Although the public tends to believe that CNN is extremely liberal, it is more reasonable to consider it as centrist biased towards left-side. People's opinion on news agencies' tendencies might be polarized. Besides, although there are significantly more famous news agencies on the liberal side, those right-leaning ones tend to support their side more firmly.

\subsubsection{Geography Distribution}
Consider results from the largest dataset (\textbf{P+all}), and with predictions coming out from \textbf{TIMME-hierarchical}.
We predict each Twitter user's ideology as either liberal or conservative. Then we calculate the percentage of the users on both sides, and depict it in Figure \ref{fig:all_states}. Darkest red represents $p \in [0, \frac{1}{8}]$ of users in that area are liberal, remaining $[\frac{7}{8}, 1]$ are conservative; darkest blue areas have $[\frac{7}{8}, 1]$ users being liberal, $[0, \frac{1}{8}]$ conservative. The intermediate colors represent the evenly-divided ranges in between. The users' locations are collected from the public information in their account profile. From our observation, conservative people are typically under-represented. \footnote{National General Election Polls data partly available at \url{https://www.realclearpolitics.com/epolls/2020/president/National.html}.}\footnote{Compare with the visualization of previous election at \url{https://en.wikipedia.org/wiki/Political_party_strength_in_U.S._states}.} For instance, as a well-known firmly-conservative state, Utah (UT) is only shown as slightly right-leaning on our map.

This is intuitively reasonable, since Twitter users are also biased. Typically biased towards youngsters and urban citizens. Although we are able to solve the problem of silent-majority by utilizing their link relations instead of text expressions, we know nothing about offline ideology. We suppose that some areas are silent on Twitter, and this guess is supported by the county-level results at Florida, shown in Figure \ref{fig:florida}. This time the color-code represents evenly-divided seven ranges from $[0, \frac{1}{7}]$ to $[\frac{6}{7}, 1]$, because of the necessity of reserving one color for representing silent areas (denoted as white for \textit{N/A}). The silent counties, typically some rural areas, have no user in our dataset, inferring that people living there do not use Twitter very often. The remaining parts of the graph makes complete sense, demonstrating a typical swing state. \footnote{The ground-truth election outcome in Florida at 2016 is at \url{https://en.wikipedia.org/wiki/2016_United_States_presidential_election_in_Florida}.}

\subsubsection{Correlated Relations}

When we train \textbf{TIMME-single} with only one relation type, some other relations' predictions benefit from it, and are becoming more and more accurate. We assume that, if by training on relation $r_i$ we achieve a good performance on relation $r_j$, then we say relation $r_i$ probably leads to $r_j$. As is shown in Figure \ref{fig:relation_relation}, relations among politicians are relatively independent except that all other relations might stimulate \textit{like}. In more ordinary user groups, \textit{reply} is the one that significantly benefit from all other relations. It is also interesting to observe that the highly-political P50 shows that \textit{like} leads to \textit{retweet}, while from more ordinary users' perspective once they \textit{liked} they are less likely to \textit{retweet}. The relations among the relations are asymmetric.

\subsubsection{Relation's Contributions to Ideology Detection}

\begin{figure}
\centering
\includegraphics[width=1.0\linewidth]{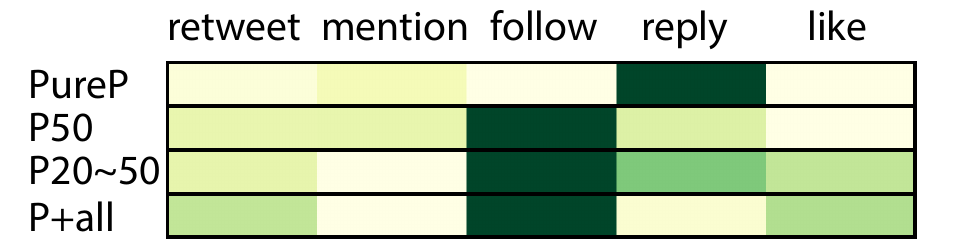}
\caption{ Illustration of $\lambda$ value in decoder on each dataset.}
\label{fig:relation_importance}
\vspace{-3mm}
\end{figure}

The importance of each relation to ideology prediction could be measured by the value of the corresponding $\lambda_r$ values in the decoder of \textbf{TIMME-hierarchical}. All the values are close to $0.2$ in practice, in $[0.99, 2.01]$, but still has some common trends, as is shown in Figure \ref{fig:relation_importance}. Despite that \textit{reply} pops out rather than \textit{follow} on \textbf{PureP}, we still insist that \textit{\textbf{follow} is the most important relation}. That is because we only crawled the most recent about $5000$ followers / followees. If a follow happened long time ago, we would not capture it. The \textit{follow} relation is especially incomplete on \textbf{PureP}.
\section{Conclusion}\label{sec:conclusion}
The \textbf{TIMME models} we proposed handles multiple relations, with a multi-relational encoder, and multi-task decoder. We step aside the silent-majority problem by relying mostly on the relations, instead of the text information.
Optionally, we accept incomplete input features, but we showed that links are able to do well on generating the ideology embedding without additional text information. From our observation, links help much more than naively-processed text in ideology-detection problem, and \textit{follow} is the most important relation to ideology detection. 
We also concluded from visualizing the state-level overall ideology map that conservative voices tend to be under-represented on Twitter. Meanwhile we confirmed that public opinions on news agencies' ideology could be polarized, with very obvious tendencies.
Our model could be easily extended to any other social network embedding problem, such as on any other dataset like Facebook as long as the dataset is legally available, and of course it works on predicting other tendencies like preferring Superman or Batman. We also believe that our dataset would be beneficial to the community.
\section{Acknowledgement}

This work is partially supported by NSF III-1705169, NSF CAREER Award 1741634, NSF \#1937599, DARPA HR00112090027, Okawa Foundation Grant, and Amazon Research Award.

Weiping Song is supported by National Key Research and Development Program of China with Grant No. 2018AAA0101900/ 2018AAA0101902 as well as the National Natural Science Foundation of China (NSFC Grant No. 61772039 and No. 91646202).

At the early stage of this work, Haoran Wang \footnote{\textit{wanghran@gmail.com}, currently working at Snap Inc.} contributed a lot to a nicely-implemented first version of the model, benefiting the rest of our work. Meanwhile, Zhiwen Hu \footnote{\textit{huzhiwen@g.ucla.edu}, currently working at PayPal.} explored the related methods' efficiencies, and his works shed light on our way.

Our team also received some external help from Yupeng Gu. He offered us his crawler code and his old dataset as references.

\bibliographystyle{ACM-Reference-Format}

\bibliography{reference}


\begin{thebibliography}{43}


\ifx \showCODEN    \undefined \def \showCODEN     #1{\unskip}     \fi
\ifx \showDOI      \undefined \def \showDOI       #1{#1}\fi
\ifx \showISBNx    \undefined \def \showISBNx     #1{\unskip}     \fi
\ifx \showISBNxiii \undefined \def \showISBNxiii  #1{\unskip}     \fi
\ifx \showISSN     \undefined \def \showISSN      #1{\unskip}     \fi
\ifx \showLCCN     \undefined \def \showLCCN      #1{\unskip}     \fi
\ifx \shownote     \undefined \def \shownote      #1{#1}          \fi
\ifx \showarticletitle \undefined \def \showarticletitle #1{#1}   \fi
\ifx \showURL      \undefined \def \showURL       {\relax}        \fi
\providecommand\bibfield[2]{#2}
\providecommand\bibinfo[2]{#2}
\providecommand\natexlab[1]{#1}
\providecommand\showeprint[2][]{arXiv:#2}

\bibitem[\protect\citeauthoryear{Achen}{Achen}{1975}]%
        {achen1975mass}
\bibfield{author}{\bibinfo{person}{Christopher~H Achen}.}
  \bibinfo{year}{1975}\natexlab{}.
\newblock \showarticletitle{Mass political attitudes and the survey response}.
\newblock \bibinfo{journal}{\emph{American Political Science Review}}
  \bibinfo{volume}{69}, \bibinfo{number}{4} (\bibinfo{year}{1975}),
  \bibinfo{pages}{1218--1231}.
\newblock


\bibitem[\protect\citeauthoryear{Baly, Karadzhov, Saleh, Glass, and Nakov}{Baly
  et~al\mbox{.}}{2019}]%
        {baly2019multi}
\bibfield{author}{\bibinfo{person}{Ramy Baly}, \bibinfo{person}{Georgi
  Karadzhov}, \bibinfo{person}{Abdelrhman Saleh}, \bibinfo{person}{James
  Glass}, {and} \bibinfo{person}{Preslav Nakov}.}
  \bibinfo{year}{2019}\natexlab{}.
\newblock \showarticletitle{Multi-task ordinal regression for jointly
  predicting the trustworthiness and the leading political ideology of news
  media}.
\newblock \bibinfo{journal}{\emph{arXiv preprint arXiv:1904.00542}}
  (\bibinfo{year}{2019}).
\newblock


\bibitem[\protect\citeauthoryear{Cen, Zou, Zhang, Yang, Zhou, and Tang}{Cen
  et~al\mbox{.}}{2019}]%
        {cen2019representation}
\bibfield{author}{\bibinfo{person}{Yukuo Cen}, \bibinfo{person}{Xu Zou},
  \bibinfo{person}{Jianwei Zhang}, \bibinfo{person}{Hongxia Yang},
  \bibinfo{person}{Jingren Zhou}, {and} \bibinfo{person}{Jie Tang}.}
  \bibinfo{year}{2019}\natexlab{}.
\newblock \showarticletitle{Representation learning for attributed multiplex
  heterogeneous network}. In \bibinfo{booktitle}{\emph{Proceedings of the 25th
  ACM SIGKDD International Conference on Knowledge Discovery \& Data Mining}}.
  \bibinfo{pages}{1358--1368}.
\newblock


\bibitem[\protect\citeauthoryear{Chen, Ma, and Xiao}{Chen
  et~al\mbox{.}}{2018}]%
        {chen2018fastgcn}
\bibfield{author}{\bibinfo{person}{Jie Chen}, \bibinfo{person}{Tengfei Ma},
  {and} \bibinfo{person}{Cao Xiao}.} \bibinfo{year}{2018}\natexlab{}.
\newblock \showarticletitle{Fastgcn: fast learning with graph convolutional
  networks via importance sampling}.
\newblock \bibinfo{journal}{\emph{arXiv preprint arXiv:1801.10247}}
  (\bibinfo{year}{2018}).
\newblock


\bibitem[\protect\citeauthoryear{Chen, Zhang, Wang, Yang, and Li}{Chen
  et~al\mbox{.}}{2017}]%
        {chen2017opinion}
\bibfield{author}{\bibinfo{person}{Wei Chen}, \bibinfo{person}{Xiao Zhang},
  \bibinfo{person}{Tengjiao Wang}, \bibinfo{person}{Bishan Yang}, {and}
  \bibinfo{person}{Yi Li}.} \bibinfo{year}{2017}\natexlab{}.
\newblock \showarticletitle{Opinion-aware Knowledge Graph for Political
  Ideology Detection.}. In \bibinfo{booktitle}{\emph{IJCAI}}.
  \bibinfo{pages}{3647--3653}.
\newblock


\bibitem[\protect\citeauthoryear{Chiang, Liu, Si, Li, Bengio, and Hsieh}{Chiang
  et~al\mbox{.}}{2019}]%
        {chiang2019cluster}
\bibfield{author}{\bibinfo{person}{Wei-Lin Chiang}, \bibinfo{person}{Xuanqing
  Liu}, \bibinfo{person}{Si Si}, \bibinfo{person}{Yang Li},
  \bibinfo{person}{Samy Bengio}, {and} \bibinfo{person}{Cho-Jui Hsieh}.}
  \bibinfo{year}{2019}\natexlab{}.
\newblock \showarticletitle{Cluster-gcn: An efficient algorithm for training
  deep and large graph convolutional networks}. In
  \bibinfo{booktitle}{\emph{Proceedings of the 25th ACM SIGKDD International
  Conference on Knowledge Discovery \& Data Mining}}.
  \bibinfo{pages}{257--266}.
\newblock


\bibitem[\protect\citeauthoryear{Clinton, Jackman, and Rivers}{Clinton
  et~al\mbox{.}}{2004}]%
        {clinton2004statistical}
\bibfield{author}{\bibinfo{person}{Joshua Clinton}, \bibinfo{person}{Simon
  Jackman}, {and} \bibinfo{person}{Douglas Rivers}.}
  \bibinfo{year}{2004}\natexlab{}.
\newblock \showarticletitle{The statistical analysis of roll call data}.
\newblock \bibinfo{journal}{\emph{American Political Science Review}}
  \bibinfo{volume}{98}, \bibinfo{number}{2} (\bibinfo{year}{2004}),
  \bibinfo{pages}{355--370}.
\newblock


\bibitem[\protect\citeauthoryear{Conover, Gon{\c{c}}alves, Ratkiewicz,
  Flammini, and Menczer}{Conover et~al\mbox{.}}{2011a}]%
        {conover2011predicting}
\bibfield{author}{\bibinfo{person}{Michael~D Conover}, \bibinfo{person}{Bruno
  Gon{\c{c}}alves}, \bibinfo{person}{Jacob Ratkiewicz},
  \bibinfo{person}{Alessandro Flammini}, {and} \bibinfo{person}{Filippo
  Menczer}.} \bibinfo{year}{2011}\natexlab{a}.
\newblock \showarticletitle{Predicting the political alignment of twitter
  users}. In \bibinfo{booktitle}{\emph{2011 IEEE third international conference
  on privacy, security, risk and trust and 2011 IEEE third international
  conference on social computing}}. IEEE, \bibinfo{pages}{192--199}.
\newblock


\bibitem[\protect\citeauthoryear{Conover, Ratkiewicz, Francisco,
  Gon{\c{c}}alves, Menczer, and Flammini}{Conover et~al\mbox{.}}{2011b}]%
        {conover2011political}
\bibfield{author}{\bibinfo{person}{Michael~D Conover}, \bibinfo{person}{Jacob
  Ratkiewicz}, \bibinfo{person}{Matthew Francisco}, \bibinfo{person}{Bruno
  Gon{\c{c}}alves}, \bibinfo{person}{Filippo Menczer}, {and}
  \bibinfo{person}{Alessandro Flammini}.} \bibinfo{year}{2011}\natexlab{b}.
\newblock \showarticletitle{Political polarization on twitter}. In
  \bibinfo{booktitle}{\emph{Fifth international AAAI conference on weblogs and
  social media}}.
\newblock


\bibitem[\protect\citeauthoryear{Culotta, Kumar, and Cutler}{Culotta
  et~al\mbox{.}}{2015}]%
        {culotta2015predicting}
\bibfield{author}{\bibinfo{person}{Aron Culotta}, \bibinfo{person}{Nirmal~Ravi
  Kumar}, {and} \bibinfo{person}{Jennifer Cutler}.}
  \bibinfo{year}{2015}\natexlab{}.
\newblock \showarticletitle{Predicting the Demographics of Twitter Users from
  Website Traffic Data.}. In \bibinfo{booktitle}{\emph{AAAI}},
  Vol.~\bibinfo{volume}{15}. Austin, TX, \bibinfo{pages}{72--8}.
\newblock


\bibitem[\protect\citeauthoryear{Defferrard, Bresson, and
  Vandergheynst}{Defferrard et~al\mbox{.}}{2016}]%
        {defferrard2016convolutional}
\bibfield{author}{\bibinfo{person}{Micha{\"e}l Defferrard},
  \bibinfo{person}{Xavier Bresson}, {and} \bibinfo{person}{Pierre
  Vandergheynst}.} \bibinfo{year}{2016}\natexlab{}.
\newblock \showarticletitle{Convolutional neural networks on graphs with fast
  localized spectral filtering}. In \bibinfo{booktitle}{\emph{Advances in
  neural information processing systems}}. \bibinfo{pages}{3844--3852}.
\newblock


\bibitem[\protect\citeauthoryear{Grover and Leskovec}{Grover and
  Leskovec}{2016}]%
        {grover2016node2vec}
\bibfield{author}{\bibinfo{person}{Aditya Grover} {and} \bibinfo{person}{Jure
  Leskovec}.} \bibinfo{year}{2016}\natexlab{}.
\newblock \showarticletitle{node2vec: Scalable feature learning for networks}.
  In \bibinfo{booktitle}{\emph{Proceedings of the 22nd ACM SIGKDD international
  conference on Knowledge discovery and data mining}}.
  \bibinfo{pages}{855--864}.
\newblock


\bibitem[\protect\citeauthoryear{Gu, Chen, Sun, and Wang}{Gu
  et~al\mbox{.}}{2016}]%
        {gu2016ideology}
\bibfield{author}{\bibinfo{person}{Yupeng Gu}, \bibinfo{person}{Ting Chen},
  \bibinfo{person}{Yizhou Sun}, {and} \bibinfo{person}{Bingyu Wang}.}
  \bibinfo{year}{2016}\natexlab{}.
\newblock \showarticletitle{Ideology detection for twitter users with
  heterogeneous types of links}.
\newblock \bibinfo{journal}{\emph{arXiv preprint arXiv:1612.08207}}
  (\bibinfo{year}{2016}).
\newblock


\bibitem[\protect\citeauthoryear{Hamilton, Ying, and Leskovec}{Hamilton
  et~al\mbox{.}}{2017}]%
        {hamilton2017inductive}
\bibfield{author}{\bibinfo{person}{Will Hamilton}, \bibinfo{person}{Zhitao
  Ying}, {and} \bibinfo{person}{Jure Leskovec}.}
  \bibinfo{year}{2017}\natexlab{}.
\newblock \showarticletitle{Inductive representation learning on large graphs}.
  In \bibinfo{booktitle}{\emph{Advances in neural information processing
  systems}}. \bibinfo{pages}{1024--1034}.
\newblock


\bibitem[\protect\citeauthoryear{Iyyer, Enns, Boyd-Graber, and Resnik}{Iyyer
  et~al\mbox{.}}{2014}]%
        {iyyer2014political}
\bibfield{author}{\bibinfo{person}{Mohit Iyyer}, \bibinfo{person}{Peter Enns},
  \bibinfo{person}{Jordan Boyd-Graber}, {and} \bibinfo{person}{Philip Resnik}.}
  \bibinfo{year}{2014}\natexlab{}.
\newblock \showarticletitle{Political ideology detection using recursive neural
  networks}. In \bibinfo{booktitle}{\emph{Proceedings of the 52nd Annual
  Meeting of the Association for Computational Linguistics (Volume 1: Long
  Papers)}}. \bibinfo{pages}{1113--1122}.
\newblock


\bibitem[\protect\citeauthoryear{Johnson and Goldwasser}{Johnson and
  Goldwasser}{2016}]%
        {johnson2016identifying}
\bibfield{author}{\bibinfo{person}{Kristen Johnson} {and} \bibinfo{person}{Dan
  Goldwasser}.} \bibinfo{year}{2016}\natexlab{}.
\newblock \showarticletitle{Identifying stance by analyzing political discourse
  on twitter}. In \bibinfo{booktitle}{\emph{Proceedings of the First Workshop
  on NLP and Computational Social Science}}. \bibinfo{pages}{66--75}.
\newblock


\bibitem[\protect\citeauthoryear{Kannangara}{Kannangara}{2018}]%
        {kannangara2018mining}
\bibfield{author}{\bibinfo{person}{Sandeepa Kannangara}.}
  \bibinfo{year}{2018}\natexlab{}.
\newblock \showarticletitle{Mining twitter for fine-grained political opinion
  polarity classification, ideology detection and sarcasm detection}. In
  \bibinfo{booktitle}{\emph{Proceedings of the Eleventh ACM International
  Conference on Web Search and Data Mining}}. \bibinfo{pages}{751--752}.
\newblock


\bibitem[\protect\citeauthoryear{Kendall, Gal, and Cipolla}{Kendall
  et~al\mbox{.}}{2018}]%
        {kendall2018multi}
\bibfield{author}{\bibinfo{person}{Alex Kendall}, \bibinfo{person}{Yarin Gal},
  {and} \bibinfo{person}{Roberto Cipolla}.} \bibinfo{year}{2018}\natexlab{}.
\newblock \showarticletitle{Multi-task learning using uncertainty to weigh
  losses for scene geometry and semantics}. In
  \bibinfo{booktitle}{\emph{Proceedings of the IEEE conference on computer
  vision and pattern recognition}}. \bibinfo{pages}{7482--7491}.
\newblock


\bibitem[\protect\citeauthoryear{Kipf and Welling}{Kipf and Welling}{2016}]%
        {kipf2016semi}
\bibfield{author}{\bibinfo{person}{Thomas~N Kipf} {and} \bibinfo{person}{Max
  Welling}.} \bibinfo{year}{2016}\natexlab{}.
\newblock \showarticletitle{Semi-supervised classification with graph
  convolutional networks}.
\newblock \bibinfo{journal}{\emph{arXiv preprint arXiv:1609.02907}}
  (\bibinfo{year}{2016}).
\newblock


\bibitem[\protect\citeauthoryear{Kuhn and Kamm}{Kuhn and Kamm}{2019}]%
        {kuhn2019national}
\bibfield{author}{\bibinfo{person}{Theresa Kuhn} {and} \bibinfo{person}{Aaron
  Kamm}.} \bibinfo{year}{2019}\natexlab{}.
\newblock \showarticletitle{The national boundaries of solidarity: a survey
  experiment on solidarity with unemployed people in the European Union}.
\newblock \bibinfo{journal}{\emph{European Political Science Review}}
  \bibinfo{volume}{11}, \bibinfo{number}{2} (\bibinfo{year}{2019}),
  \bibinfo{pages}{179--195}.
\newblock


\bibitem[\protect\citeauthoryear{Li, Han, and Wu}{Li et~al\mbox{.}}{2018}]%
        {li2018deeper}
\bibfield{author}{\bibinfo{person}{Qimai Li}, \bibinfo{person}{Zhichao Han},
  {and} \bibinfo{person}{Xiao-Ming Wu}.} \bibinfo{year}{2018}\natexlab{}.
\newblock \showarticletitle{Deeper insights into graph convolutional networks
  for semi-supervised learning}. In \bibinfo{booktitle}{\emph{Thirty-Second
  AAAI Conference on Artificial Intelligence}}.
\newblock


\bibitem[\protect\citeauthoryear{Liu, Chen, Yang, Zhou, Li, and Song}{Liu
  et~al\mbox{.}}{2018}]%
        {liu2018heterogeneous}
\bibfield{author}{\bibinfo{person}{Ziqi Liu}, \bibinfo{person}{Chaochao Chen},
  \bibinfo{person}{Xinxing Yang}, \bibinfo{person}{Jun Zhou},
  \bibinfo{person}{Xiaolong Li}, {and} \bibinfo{person}{Le Song}.}
  \bibinfo{year}{2018}\natexlab{}.
\newblock \showarticletitle{Heterogeneous graph neural networks for malicious
  account detection}. In \bibinfo{booktitle}{\emph{Proceedings of the 27th ACM
  International Conference on Information and Knowledge Management}}.
  \bibinfo{pages}{2077--2085}.
\newblock


\bibitem[\protect\citeauthoryear{Martini and Torcal}{Martini and
  Torcal}{2019}]%
        {martini2019trust}
\bibfield{author}{\bibinfo{person}{Sergio Martini} {and}
  \bibinfo{person}{Mariano Torcal}.} \bibinfo{year}{2019}\natexlab{}.
\newblock \showarticletitle{Trust across political conflicts: Evidence from a
  survey experiment in divided societies}.
\newblock \bibinfo{journal}{\emph{Party Politics}} \bibinfo{volume}{25},
  \bibinfo{number}{2} (\bibinfo{year}{2019}), \bibinfo{pages}{126--139}.
\newblock


\bibitem[\protect\citeauthoryear{Nguyen, Boyd-Graber, Resnik, and Miler}{Nguyen
  et~al\mbox{.}}{2015}]%
        {nguyen2015tea}
\bibfield{author}{\bibinfo{person}{Viet-An Nguyen}, \bibinfo{person}{Jordan
  Boyd-Graber}, \bibinfo{person}{Philip Resnik}, {and}
  \bibinfo{person}{Kristina Miler}.} \bibinfo{year}{2015}\natexlab{}.
\newblock \showarticletitle{Tea party in the house: A hierarchical ideal point
  topic model and its application to republican legislators in the 112th
  congress}. In \bibinfo{booktitle}{\emph{Proceedings of the 53rd Annual
  Meeting of the Association for Computational Linguistics and the 7th
  International Joint Conference on Natural Language Processing (Volume 1: Long
  Papers)}}. \bibinfo{pages}{1438--1448}.
\newblock


\bibitem[\protect\citeauthoryear{Park}{Park}{2013}]%
        {park2013does}
\bibfield{author}{\bibinfo{person}{Chang~Sup Park}.}
  \bibinfo{year}{2013}\natexlab{}.
\newblock \showarticletitle{Does Twitter motivate involvement in politics?
  Tweeting, opinion leadership, and political engagement}.
\newblock \bibinfo{journal}{\emph{Computers in Human Behavior}}
  \bibinfo{volume}{29}, \bibinfo{number}{4} (\bibinfo{year}{2013}),
  \bibinfo{pages}{1641--1648}.
\newblock


\bibitem[\protect\citeauthoryear{Pennington, Socher, and Manning}{Pennington
  et~al\mbox{.}}{2014}]%
        {pennington2014glove}
\bibfield{author}{\bibinfo{person}{Jeffrey Pennington},
  \bibinfo{person}{Richard Socher}, {and} \bibinfo{person}{Christopher~D
  Manning}.} \bibinfo{year}{2014}\natexlab{}.
\newblock \showarticletitle{Glove: Global vectors for word representation}. In
  \bibinfo{booktitle}{\emph{Proceedings of the 2014 conference on empirical
  methods in natural language processing (EMNLP)}}.
  \bibinfo{pages}{1532--1543}.
\newblock


\bibitem[\protect\citeauthoryear{Pollock, Brock, and Ellison}{Pollock
  et~al\mbox{.}}{2015}]%
        {pollock2015populism}
\bibfield{author}{\bibinfo{person}{Gary Pollock}, \bibinfo{person}{Tom Brock},
  {and} \bibinfo{person}{Mark Ellison}.} \bibinfo{year}{2015}\natexlab{}.
\newblock \showarticletitle{Populism, ideology and contradiction: mapping young
  people's political views}.
\newblock \bibinfo{journal}{\emph{The Sociological Review}}
  \bibinfo{volume}{63} (\bibinfo{year}{2015}), \bibinfo{pages}{141--166}.
\newblock


\bibitem[\protect\citeauthoryear{Poole and Rosenthal}{Poole and
  Rosenthal}{1985}]%
        {poole1985spatial}
\bibfield{author}{\bibinfo{person}{Keith~T Poole} {and} \bibinfo{person}{Howard
  Rosenthal}.} \bibinfo{year}{1985}\natexlab{}.
\newblock \showarticletitle{A spatial model for legislative roll call
  analysis}.
\newblock \bibinfo{journal}{\emph{American Journal of Political Science}}
  (\bibinfo{year}{1985}), \bibinfo{pages}{357--384}.
\newblock


\bibitem[\protect\citeauthoryear{Preo{\c{t}}iuc-Pietro, Liu, Hopkins, and
  Ungar}{Preo{\c{t}}iuc-Pietro et~al\mbox{.}}{2017}]%
        {preoctiuc2017beyond}
\bibfield{author}{\bibinfo{person}{Daniel Preo{\c{t}}iuc-Pietro},
  \bibinfo{person}{Ye Liu}, \bibinfo{person}{Daniel Hopkins}, {and}
  \bibinfo{person}{Lyle Ungar}.} \bibinfo{year}{2017}\natexlab{}.
\newblock \showarticletitle{Beyond binary labels: political ideology prediction
  of twitter users}. In \bibinfo{booktitle}{\emph{Proceedings of the 55th
  Annual Meeting of the Association for Computational Linguistics (Volume 1:
  Long Papers)}}. \bibinfo{pages}{729--740}.
\newblock


\bibitem[\protect\citeauthoryear{Reimers and Gurevych}{Reimers and
  Gurevych}{2019}]%
        {reimers2019sentence}
\bibfield{author}{\bibinfo{person}{Nils Reimers} {and} \bibinfo{person}{Iryna
  Gurevych}.} \bibinfo{year}{2019}\natexlab{}.
\newblock \showarticletitle{Sentence-bert: Sentence embeddings using siamese
  bert-networks}.
\newblock \bibinfo{journal}{\emph{arXiv preprint arXiv:1908.10084}}
  (\bibinfo{year}{2019}).
\newblock


\bibitem[\protect\citeauthoryear{Ruder}{Ruder}{2017}]%
        {ruder2017overview}
\bibfield{author}{\bibinfo{person}{Sebastian Ruder}.}
  \bibinfo{year}{2017}\natexlab{}.
\newblock \showarticletitle{An overview of multi-task learning in deep neural
  networks}.
\newblock \bibinfo{journal}{\emph{arXiv preprint arXiv:1706.05098}}
  (\bibinfo{year}{2017}).
\newblock


\bibitem[\protect\citeauthoryear{Schlichtkrull, Kipf, Bloem, Van Den~Berg,
  Titov, and Welling}{Schlichtkrull et~al\mbox{.}}{2018}]%
        {schlichtkrull2018modeling}
\bibfield{author}{\bibinfo{person}{Michael Schlichtkrull},
  \bibinfo{person}{Thomas~N Kipf}, \bibinfo{person}{Peter Bloem},
  \bibinfo{person}{Rianne Van Den~Berg}, \bibinfo{person}{Ivan Titov}, {and}
  \bibinfo{person}{Max Welling}.} \bibinfo{year}{2018}\natexlab{}.
\newblock \showarticletitle{Modeling relational data with graph convolutional
  networks}. In \bibinfo{booktitle}{\emph{European Semantic Web Conference}}.
  Springer, \bibinfo{pages}{593--607}.
\newblock


\bibitem[\protect\citeauthoryear{Socher, Chen, Manning, and Ng}{Socher
  et~al\mbox{.}}{2013}]%
        {socher2013reasoning}
\bibfield{author}{\bibinfo{person}{Richard Socher}, \bibinfo{person}{Danqi
  Chen}, \bibinfo{person}{Christopher~D Manning}, {and} \bibinfo{person}{Andrew
  Ng}.} \bibinfo{year}{2013}\natexlab{}.
\newblock \showarticletitle{Reasoning with neural tensor networks for knowledge
  base completion}. In \bibinfo{booktitle}{\emph{Advances in neural information
  processing systems}}. \bibinfo{pages}{926--934}.
\newblock


\bibitem[\protect\citeauthoryear{Sun and Han}{Sun and Han}{2012}]%
        {sun2012mining}
\bibfield{author}{\bibinfo{person}{Yizhou Sun} {and} \bibinfo{person}{Jiawei
  Han}.} \bibinfo{year}{2012}\natexlab{}.
\newblock \showarticletitle{Mining heterogeneous information networks:
  principles and methodologies}.
\newblock \bibinfo{journal}{\emph{Synthesis Lectures on Data Mining and
  Knowledge Discovery}} \bibinfo{volume}{3}, \bibinfo{number}{2}
  (\bibinfo{year}{2012}), \bibinfo{pages}{1--159}.
\newblock


\bibitem[\protect\citeauthoryear{Tang, Qu, Wang, Zhang, Yan, and Mei}{Tang
  et~al\mbox{.}}{2015}]%
        {tang2015line}
\bibfield{author}{\bibinfo{person}{Jian Tang}, \bibinfo{person}{Meng Qu},
  \bibinfo{person}{Mingzhe Wang}, \bibinfo{person}{Ming Zhang},
  \bibinfo{person}{Jun Yan}, {and} \bibinfo{person}{Qiaozhu Mei}.}
  \bibinfo{year}{2015}\natexlab{}.
\newblock \showarticletitle{Line: Large-scale information network embedding}.
  In \bibinfo{booktitle}{\emph{Proceedings of the 24th international conference
  on world wide web}}. \bibinfo{pages}{1067--1077}.
\newblock


\bibitem[\protect\citeauthoryear{Veli{\v{c}}kovi{\'c}, Cucurull, Casanova,
  Romero, Lio, and Bengio}{Veli{\v{c}}kovi{\'c} et~al\mbox{.}}{2017}]%
        {velivckovic2017graph}
\bibfield{author}{\bibinfo{person}{Petar Veli{\v{c}}kovi{\'c}},
  \bibinfo{person}{Guillem Cucurull}, \bibinfo{person}{Arantxa Casanova},
  \bibinfo{person}{Adriana Romero}, \bibinfo{person}{Pietro Lio}, {and}
  \bibinfo{person}{Yoshua Bengio}.} \bibinfo{year}{2017}\natexlab{}.
\newblock \showarticletitle{Graph attention networks}.
\newblock \bibinfo{journal}{\emph{arXiv preprint arXiv:1710.10903}}
  (\bibinfo{year}{2017}).
\newblock


\bibitem[\protect\citeauthoryear{Vijayaraghavan, Vosoughi, and
  Roy}{Vijayaraghavan et~al\mbox{.}}{2017}]%
        {vijayaraghavan2017twitter}
\bibfield{author}{\bibinfo{person}{Prashanth Vijayaraghavan},
  \bibinfo{person}{Soroush Vosoughi}, {and} \bibinfo{person}{Deb Roy}.}
  \bibinfo{year}{2017}\natexlab{}.
\newblock \showarticletitle{Twitter demographic classification using deep
  multi-modal multi-task learning}. In \bibinfo{booktitle}{\emph{Proceedings of
  the 55th Annual Meeting of the Association for Computational Linguistics
  (Volume 2: Short Papers)}}. \bibinfo{pages}{478--483}.
\newblock


\bibitem[\protect\citeauthoryear{Wang, Zhang, Hou, Xie, Guo, and Liu}{Wang
  et~al\mbox{.}}{2018}]%
        {wang2018shine}
\bibfield{author}{\bibinfo{person}{Hongwei Wang}, \bibinfo{person}{Fuzheng
  Zhang}, \bibinfo{person}{Min Hou}, \bibinfo{person}{Xing Xie},
  \bibinfo{person}{Minyi Guo}, {and} \bibinfo{person}{Qi Liu}.}
  \bibinfo{year}{2018}\natexlab{}.
\newblock \showarticletitle{Shine: Signed heterogeneous information network
  embedding for sentiment link prediction}. In
  \bibinfo{booktitle}{\emph{Proceedings of the Eleventh ACM International
  Conference on Web Search and Data Mining}}. \bibinfo{pages}{592--600}.
\newblock


\bibitem[\protect\citeauthoryear{Wang, Ji, Shi, Wang, Ye, Cui, and Yu}{Wang
  et~al\mbox{.}}{2019}]%
        {wang2019heterogeneous}
\bibfield{author}{\bibinfo{person}{Xiao Wang}, \bibinfo{person}{Houye Ji},
  \bibinfo{person}{Chuan Shi}, \bibinfo{person}{Bai Wang},
  \bibinfo{person}{Yanfang Ye}, \bibinfo{person}{Peng Cui}, {and}
  \bibinfo{person}{Philip~S Yu}.} \bibinfo{year}{2019}\natexlab{}.
\newblock \showarticletitle{Heterogeneous graph attention network}. In
  \bibinfo{booktitle}{\emph{The World Wide Web Conference}}.
  \bibinfo{pages}{2022--2032}.
\newblock


\bibitem[\protect\citeauthoryear{Xu, Hu, Leskovec, and Jegelka}{Xu
  et~al\mbox{.}}{2018}]%
        {xu2018powerful}
\bibfield{author}{\bibinfo{person}{Keyulu Xu}, \bibinfo{person}{Weihua Hu},
  \bibinfo{person}{Jure Leskovec}, {and} \bibinfo{person}{Stefanie Jegelka}.}
  \bibinfo{year}{2018}\natexlab{}.
\newblock \showarticletitle{How powerful are graph neural networks?}
\newblock \bibinfo{journal}{\emph{arXiv preprint arXiv:1810.00826}}
  (\bibinfo{year}{2018}).
\newblock


\bibitem[\protect\citeauthoryear{Yang, Yih, He, Gao, and Deng}{Yang
  et~al\mbox{.}}{2014}]%
        {yang2014embedding}
\bibfield{author}{\bibinfo{person}{Bishan Yang}, \bibinfo{person}{Wen-tau Yih},
  \bibinfo{person}{Xiaodong He}, \bibinfo{person}{Jianfeng Gao}, {and}
  \bibinfo{person}{Li Deng}.} \bibinfo{year}{2014}\natexlab{}.
\newblock \showarticletitle{Embedding entities and relations for learning and
  inference in knowledge bases}.
\newblock \bibinfo{journal}{\emph{arXiv preprint arXiv:1412.6575}}
  (\bibinfo{year}{2014}).
\newblock


\bibitem[\protect\citeauthoryear{Yun, Jeong, Kim, Kang, and Kim}{Yun
  et~al\mbox{.}}{2019}]%
        {yun2019graph}
\bibfield{author}{\bibinfo{person}{Seongjun Yun}, \bibinfo{person}{Minbyul
  Jeong}, \bibinfo{person}{Raehyun Kim}, \bibinfo{person}{Jaewoo Kang}, {and}
  \bibinfo{person}{Hyunwoo~J Kim}.} \bibinfo{year}{2019}\natexlab{}.
\newblock \showarticletitle{Graph Transformer Networks}. In
  \bibinfo{booktitle}{\emph{Advances in Neural Information Processing
  Systems}}. \bibinfo{pages}{11960--11970}.
\newblock


\bibitem[\protect\citeauthoryear{Zhang, Song, Huang, Swami, and Chawla}{Zhang
  et~al\mbox{.}}{2019}]%
        {zhang2019heterogeneous}
\bibfield{author}{\bibinfo{person}{Chuxu Zhang}, \bibinfo{person}{Dongjin
  Song}, \bibinfo{person}{Chao Huang}, \bibinfo{person}{Ananthram Swami}, {and}
  \bibinfo{person}{Nitesh~V Chawla}.} \bibinfo{year}{2019}\natexlab{}.
\newblock \showarticletitle{Heterogeneous graph neural network}. In
  \bibinfo{booktitle}{\emph{Proceedings of the 25th ACM SIGKDD International
  Conference on Knowledge Discovery \& Data Mining}}.
  \bibinfo{pages}{793--803}.
\newblock


\end{thebibliography}

\clearpage

\appendix
\section{Data Preparation}\label{append_sec:dataprep_details}

We target at building a dataset representing the political-centered social network (Section \ref{sec:problem}), a selected subset from the giant Twitter network. Handling this dataset would be challenging. For example, for GraphSAGE, neighborhood-sampling can not be easily done both effectively and efficiently. Our dataset reaches the blind spots of many existing models. 

The tools we used to crawl politicians' name lists from the government website, and their potential Twitter accounts from Google, is Scrapy. \footnote{\url{https://scrapy.org/}} To legally and reliably crawl from Twitter data, we first applied for Developer API from Twitter \footnote{\url{https://developer.twitter.com/}}, and then used Tweepy \footnote{\url{https://www.tweepy.org/}} for crawling. We set very strict rate limits for our crawlers so as not to harm any server. 
Our dataset is released at \url{https://github.com/PatriciaXiao/TIMME}.
Raw data was collected by April, 2019.

\subsection{Twitter IDs Preparation}

Let us take the same notation as in Section \ref{sec:problem}, describing the process as: to construct $\mathcal{G}_p = \{ \mathcal{V}, \{ \mathcal{E}_1, \mathcal{E}_2, \mathcal{E}_3, \mathcal{E}_4, \mathcal{E}_5 \} \}$, we first select the users to be included $\mathcal{V}$, then we include the links among vertices in $\mathcal{V}$ under each relation $r \in \mathcal{R} = \{ 1, 2, 3, 4, 5 \}$ into $\mathcal{E}_r$ accordingly.

\subsubsection{Politicians Twitter IDs}

As is described briefly in Section \ref{subsec:dataprep}, we need to start from a set of politicians $\mathcal{P}$, which we treat as seeds for further crawling.

To start with, we first get the name list of the recently-active politicians, consists of:
\begin{itemize}[leftmargin=*]
    \item The union-set of $115^{th}$ and $116^{th}$ US congress members, where we observe a lot of overlap between the two groups; \footnote{Congress members' name list with party information is publicly available at \url{https://www.congress.gov/members} .}
    \item Recent-years' presidents and their cabinets; \footnote{Obama and Trump's cabinet is publicly available at \url{https://obamawhitehouse.archives.gov/administration/cabinet} and \url{https://www.whitehouse.gov/the-trump-administration/the-cabinet/} respectively}
    \item Additional politicians must be included: Hilary Clinton, who was running for the president of the United States not long ago; Michelle Obama, who was the former First Lady.
\end{itemize}

Next, with the help of Google, we crawled the most-likely Twitter names and IDs of the politicians. We do so automatically, by providing Google a politician's name and the keyword ``twitter'', and parsing the first response. Then after manual filtering, we have $583$ politicians' Twitter accounts available, who make up our politicians set $\mathcal{P}$. Anyone else to be included in our dataset must be in the 1-hop neighborhood of a politician (Section \ref{sec:problem}).

\subsubsection{Candidate Non-Politicians Twitter IDs}

With the help of Twitter Developer API, we are able to get the \textbf{full} followers and followees list of any Twitter user. 

However, it is not affordable to include all followers and followees of the politicians, thus we set a limit on \textit{window size} $s$ when crawling the candidate non-politicians list, only accepting the most-recent $s = 5,000$ followers or followees of any politician. These followers and followees we collected form a raw candidate set $\mathcal{C}_{raw}$. Then we remove the politicians from this set, resulting in the final candidates set $\mathcal{C} = \mathcal{C}_{raw} - \mathcal{P}$.
$\forall v_i \in \mathcal{C}$, we apply the same window size $s = 5,000$ and crawled their most recent $s$ followers, $s$ followees. All follower-followee pairs are stored into a database for the convenience of the following steps.

\subsubsection{Selecting Subgroups from Candidates}

$\mathcal{C}$ is still too large a user set, and chaotic, as we don't know anything about its components. To conduct meaningful analysis, we need to select some meaningful subgroups from it, such as a very-political subgroup, and a political-outliers subgroup, etc.

The criteria we used to select the desired subgroups of users is some thresholds. We define a political-measurement $t_i$ for each user $v_i \in \mathcal{C}$, who is followed by $t_{i,1}$ politicians $p \in \mathcal{P}$, and meanwhile following $t_{i,2}$ politicians, thus $t_i$ is computed by $t_i = \max(t_{i,1}, t_{i,2})$.

Then we set a threshold range $t$, set upon each $t_i$, used for filtering the groups of users. Considering we set $t$ as threshold range for graph $\mathcal{G}_p$, $\forall v_i \in \mathcal{V}$, if $t_i \in t$, then $v_i \in \mathcal{G}_p$, otherwise $v_i \notin \mathcal{G}_p$. By having $t = \{\infty\}$, we select a minimum subgraph containing purely politicians, resulting in our \textbf{PureP} dataset. $t \in [50, \infty)$ allows us to select a small group of users who are keen on political topics, together with the politicians, being our \textbf{P50} dataset. $t \in [20, 50)$ for less-political users, plus the politicians, being our \textbf{P20$\sim$50} dataset. $t \in [20, \infty)$ includes all nodes $v_i$ whose $t_i \geq 20$. We want to have a dataset representing more general users, containing some users from each group. Therefore, we include another $3,000$ users randomly selected from the group $t \in [1,5)$. Adding these random political-outlier users will make the dataset resembles the real network even more. Putting together the politicians, $t \in [20, \infty)$ group, and the $3,000$ random outliers from $t \in [0,5)$ group, we form the dataset \textbf{P+all}. Ideally, \textbf{P+all} has representatives of all groups of users on Twitter. The statistics are concluded in Table \ref{tab::stat}.

\subsection{Relation Preparation}

Only the \textit{follow} relation is directly observed and already well-prepared at this stage (stored in a database, as we mentioned before). Other Twitter relations: \textit{retweet}, \textit{mention}, \textit{like}, \textit{reply}, must be concluded from tweets. We distinguish the different relation types from the tweets by the tweets' fields in responded JSON from API. For example, there are some fields indicating if an \textbf{``\@''} mark is a \textit{mention}, a \textit{retweet}, or it links to nothing. According to our observation, the fields in the Json file responded from Twitter API might change across time. We don't know when will it be the next update, so there's no ground-truth solution for this part. We suggest whoever want to do so test the crawler first on her/his own account, trying all behaviors to conclude some patterns. Note: rate limit applies. \footnote{\url{https://developer.twitter.com/en/docs/basics/rate-limiting}}

Due to the Twitter official API limits, the maximum amount of tweets we could crawl for each user along the timeline is around $3,200$. Therefore, all relations are incomplete. All links we have only reflect some recent interactions among the users.

\subsection{Feature Preparation}\label{append_sec:feature}

We get feature from text, using a user's tweets posted to generate her/his feature.
Although there has been some recent advances in NLP with transformer-based structures, such as  BERT and XLNet, Sentence-BERT \cite{reimers2019sentence} found that BERT / XLNet embeddings are generally performing worse than GloVe \cite{pennington2014glove} average on sentence-level tasks. Not to mention the computational cost of transformers. We therefore use GloVe-average of the words as features, Wikipedia 2014 + Gigaword 5 (300d) pre-trained version.
When we apply the average-GloVe embedding on tweet-level, and want to tell the ideology behind the tweets, we could easily achieve $\approx 72.84\%$ accuracy, using a $2$-layers MLP, after only $200$ epochs of training.

\subsection{Label Preparation}

If we are to use only the $583$ labels from the politicians, the evaluation will always be untrustworthy. To overcome this issue, we manually expand the labels.
We first crawled the users' profiles of $\forall v_i \in \mathcal{P} \cup \mathcal{C}$, getting their information such as location and account description. Next, using the descriptions, searching for the words \textit{democratic},  \textit{republican}, \textit{conservative},  \textit{liberal}, their correct spell and variations, we have a large group of candidates. Then we do manual filtering to get rid of the uncertain users, reading their descriptions and recent tweets. We successfully included $2,976$ high-quality new labels in the end. Those labels make the node-classification task significantly more stable and reliable.

\section{Proof of Weight being Feature}\label{append_sec:proof}

Starting from our layer-wise propagation formula, we have that, at the first convolutional layer (notations in Section \ref{sec:model}):
\begin{displaymath}
H^{(1)} = \sigma\Big( \sum_{r \in \hat{\mathcal{R}}} \alpha_r \hat{A}_r H^{(0)} W_r^{(0)} \Big)
\end{displaymath}
where $H^{(0)} \in N \times d^{(0)}$ is the input feature-matrix. When using one-hot embedding of features, $H^{(0)} = I$ and $d^{(0)} = N$, thus the right-hand-side is equivalent with $\sigma\Big( \sum_{r \in \hat{\mathcal{R}}} \alpha_r \hat{A}_r W_r^{(0)} \Big)$. Now, $W_r^{(0)}$ on its own plays the role of $H^{(0)} W_r^{(0)}$ when $H^{(0)} \neq I$. Previously, relation $r$'s propagation could be viewed as aggregation of a linear transformation ($W_r^{(0)}$) done on $H^{(0)}$, from the neighborhood ($\hat{A}_r$) of each node under relation $r$. Now, it could simply be viewed as the propagation of $W_r^{(0)} \in \mathbb{R}^{N \times d^{(1)}}$. From another point of view, it is equivalent as having input features being $\tilde{H}^{(0)} = W_r^{(0)} \in \mathbb{R}^{N \times d^{(1)}}$, and set $\tilde{W}_r^{(0)} \in \mathbb{R}^{d^{(1)} \times d^{(1)}} = I_{d^{(1)}}$ being fixed identical matrix not to be updated. That's the reason why we believe that $W_r^{(0)}$ captures the nodes' learned features under relation $r$.

\section{Baseline Hyper-Parameter and Architectural Optimizations}\label{append_sec:baseline}

\subsection{Applying GCN model Directly}

As is discussed in Section~\ref{sec:ref}, due to the uniqueness of the \textit{political-centered social network dataset}, most of the existing models won't work well under our problem settings. We want to examine how well could GCN do when treating all relations as the same, ignoring the heterogeneous types. Very interestingly, without much work on hyper-parameter optimization, we only increased the hidden size and added the learning rate scheduler, it works pretty well. This phenomenon could potentially be an indirect evidence that relations are correlated, in addition to the discussions in Section \ref{sec:experiment}.

\subsection{Missing-Task Completion}
We compare our model's performance on each task with the baselines.
Ideally, we want models working on heterogeneous information networks with both node-classification task and link-prediction task as our baselines, so that we could compare with them directly. However, the situation we faced was not as easy as such. For instance, GCN and HAN never considered applying themselves directly on link-prediction tasks. But we all know that once we have the embeddings of the nodes, link prediction is doable.

Therefore, we decided that whenever a baseline originally couldn't handle a task, we lend it our decoder's task-specific cells. This decision brings about some significant improvements on the link prediction performances of NTN+ and GCN+, since \textbf{TIMME-NTN} is powerful and efficient for link-prediction. Just in case, we also decide that when a node-classification task is missing, we should add a linear transformation layer with output units $2$, the same as what we did, and apply a simple cross-entropy loss. From this perspective, it is no longer fair to compare them with r-GCN directly. To distinguish them from others' standard models, we add a plus sign \textbf{``+''} to the names, indicating that ``we lend it our cells''.

\subsection{Optimizing r-GCN}

The most important contribution of r-GCN is the weight-matrix decomposition methods. This mechanism would be very helpful in reducing the parameters, especially when the number of relations $R$ is super high. However, in our case where $R$ is small, the weight-decomposition operation is counter-effective. The first option, basis decomposition, the number of basis $b$ is easily being larger than $R$. In the second option, block-diagonal decomposition, reduces the parameter size too dramatically, and harms the model's performance. Reviewing the experiments reported in the r-GCN paper, seeing how they chose these hyper-parameters across datasets, we found that when $R$ is small, they often chose basis-decomposition with $b=0$. We go by the same option, which works well in practice.

\subsection{Optimizing HAN}
HAN/HAN+, in general, because of the complex structure with a lot of parameters, gets easily over-fitting. What makes things worse, its training curve is never stable, and our early tryouts on using validation set to automatically stop it at an optimal point did not work well. We had do it manually, by verifying when its best result appears on the validation set and when over-fitting starts, finding the right time to stop training. By default, we set learning rate $0.005$, regularization parameter $0.001$, the semantic-level attention-vector dimension $128$, multi-head-attention cell's number of heads $K=8$. We set the hyper-parameters in the \textbf{TIMME-NTN} component of HAN+ the same with ours. Optimizing HAN was a tough work to do, for it requires re-adapting every choices we made on every dataset for every task. Adding more meta-path would potentially boosting its performance, but the computational cost will be overwhelming. Another observation is that, \textbf{TIMME models} are significantly better than HAN/HAN+ in handling imperfect features. When using GloVe-average features, \textbf{TIMME models} typically perform about $1\%$ worse than using one-hot features, while HAN/HAN+ experience performance-drop up to around $10\%$.

\end{document}